\documentclass[letterpaper,conference]{ieeeconf}       
\IEEEoverridecommandlockouts                              

\usepackage{graphicx}
\usepackage{algpseudocode}
\usepackage{amsmath, amssymb, mathtools}
\usepackage{cancel}
\usepackage[most]{tcolorbox}
\tcbuselibrary{listings, breakable}
\usepackage{xcolor}
\usepackage{booktabs} 
\usepackage{multirow} 
\usepackage[caption=false, font=footnotesize]{subfig}
\usepackage{makecell}
\usepackage[normalem]{ulem}
\usepackage{flushend}
\usepackage[ruled,vlined]{algorithm2e}
\usepackage{pifont}
\usepackage{makecell}
\usepackage{siunitx}

\definecolor{codegray}{rgb}{0.5,0.5,0.5}
\definecolor{colorbackgroud}{rgb}{1,1,1}
\definecolor{borderblue}{rgb}{0.36,0.36,0.56}
\definecolor{codegreen}{rgb}{0.26,0.76,0.26}
\definecolor{codeblue}{rgb}{0.9, 0.9, 0.9}
\definecolor{longhri}{rgb}{0.6, 0.6, 0.6}
\definecolor{dmpcolor}{rgb}{0.75, 0.75, 0.75}

\newtcolorbox{codebox}{
    before = \vspace{5pt},   
    colback=colorbackgroud, 
    colframe=borderblue, 
    boxrule=0.5pt, 
    arc=0pt, 
    left=0pt, 
    right=6pt, 
    boxsep=0pt, 
    top=1pt,           
    bottom=1pt,
    breakable,
    enhanced jigsaw,
    text width=\dimexpr\columnwidth-10pt\relax
}
\newtcolorbox{codebox1}{
    before = \vspace{5pt},
    colback=colorbackgroud, 
    colframe=borderblue, 
    boxrule=0.5pt, 
    arc=0pt, 
    left=0pt, 
    right=0pt, 
    boxsep=0pt, 
    top=1pt,           
    bottom=1pt,
    breakable,
    enhanced jigsaw,
    text width=\dimexpr\textwidth-10pt\relax
}
\definecolor{mygray}{gray}{0.28}

\usepackage[hidelinks]{hyperref}
\usepackage{eso-pic,xcolor}
\newcommand{\vordoi}{\href{https://doi.org/10.1002/aisy.202500640}{10.1002/aisy.202500640}}

\newcommand{\AISacceptedTopBanner}{%
  \AddToShipoutPictureFG*{%
    \AtPageUpperLeft{%
      \raisebox{-1.2cm}[0pt][0pt]{%
        \begin{minipage}{\paperwidth}
          \centering\footnotesize
          \fboxsep=4pt
          \colorbox{black!5}{%
            \parbox{\dimexpr\paperwidth-2cm\relax}{\centering
              \textbf{Accepted \& Published in \emph{Advanced Intelligent Systems}}\\
              Version of Record (VoR): \vordoi \quad
            }%
          }
        \end{minipage}%
      }%
    }%
  }%
}

\begin{document}
\AISacceptedTopBanner

\title{Hierarchical Language Models for Semantic Navigation and Manipulation in an Aerial-Ground Robotic System}


\author{Haokun~Liu, Zhaoqi~Ma, Yunong Li, Junichiro Sugihara, Yicheng Chen, Jinjie Li, and Moju Zhao$^{\ast}$
\thanks{Haokun Liu, Zhaoqi Ma, Yunong Li, Junichiro Sugihara, Yicheng Chen, Jinjie Li, Moju Zhao are with DRAGON Lab at Department of Mechanical Engineering, The University of Tokyo, Tokyo, 113-8654, Japan (e-mail: \{haokun-liu, zhaoqi-ma, yunong-li, j-sugihara, yicheng-chen, jinjie-li, chou\}@dragon.t.u-tokyo.ac.jp. \textit{($^{\ast}$Corresponding author: Moju Zhao.)}}
}

\maketitle





\begin{abstract}

Heterogeneous multi-robot systems show great potential in complex tasks requiring coordinated hybrid cooperation. However, existing methods rely on \textcolor{black}{static or task-specific models often lack generalizability across diverse tasks and dynamic environments}. This highlights the need for generalizable intelligence that can bridge high-level reasoning with low-level execution across heterogeneous agents.
To address this, we propose a \textcolor{black}{hierarchical multimodal framework} that integrates a prompted Large Language Model (LLM) with a \textcolor{black}{fine-tuned Vision-Language Model (VLM)}.
\textcolor{black}{At the system level}, the LLM performs hierarchical task decomposition and constructs a global semantic map, \textcolor{black}{while the VLM provides semantic perception and object localization, where the proposed GridMask significantly enhances the VLM's spatial accuracy for reliable fine-grained manipulation.}
The aerial robot leverages this global map to generate semantic paths and guide the ground robot’s local navigation and manipulation, ensuring robust coordination even in target-absent or ambiguous scenarios.
We validate the framework through extensive \textcolor{black}{simulation} and real-world experiments on long-horizon object arrangement tasks, demonstrating \textcolor{black}{zero-shot adaptability, robust semantic navigation, and reliable manipulation in dynamic environments}.
To the best of our knowledge, this work presents the first heterogeneous aerial–ground robotic system that integrates VLM-based perception with LLM-driven reasoning for global high-level task planning and execution.

\end{abstract}
\IEEEpeerreviewmaketitle
\section{Introduction}

Heterogeneous Multi-Robot Systems (HMRS), composed of agents with diverse capabilities—such as aerial, ground, or underwater robots—are well-suited for complex tasks that demand a combination of manipulation, navigation, and observation \cite{yan2013survey}. Compared to homogeneous systems \cite{homo_multiagent_tro}, HMRS provides greater flexibility and efficiency in dynamic environments. However, unlocking their full potential remains a significant challenge due to the inherent complexity of cross-modal coordination. Traditional approaches to robot control and coordination often rely on pre-programming and explicit communication, which struggle to adapt to unforeseen situations—a limitation commonly observed in homogeneous systems \cite{coppola2020survey}, and further intensified in heterogeneous systems due to their cross-modal complexity and coordination requirements.

In dynamic environments, where tasks and conditions rapidly change, static models and pre-defined behaviors often lead to suboptimal decisions and poor coordination \cite{rizk2019cooperative}. Traditional control strategies struggle to generalize beyond fixed scenarios, particularly in task allocation, motion planning, and manipulation in dynamic environments. This limitation highlights the need for adaptable intelligence capable of decomposing high-level instructions for task allocation, performing motion planning, and executing manipulation tasks, without task-specific training.

This need for adaptable intelligence has been catalyzed by recent advancements in large-scale multimodal models, such as GPT \cite{openai2024gpt4} and Gemini \cite{geminiteam2024geminifamilyhighlycapable}, which have enabled Multi-Agent Language Model (MA-LLM) systems that combine high-level reasoning with visual perception. The MA-LLM frameworks utilize the semantic reasoning capabilities of LLMs and the perceptual grounding offered by VLMs, allowing robots to process both linguistic and visual information.
While LLMs can decompose high-level tasks into executable motion functions \cite{codeaspolicy}, and VLMs provide semantic and spatial information from images \cite{VLM_icra2024}, directly integrating reasoning and perception into a single-layer system often results in tight coupling. This tight coupling makes the system brittle to perception errors and creates difficulties in maintaining long-horizon tasks. To address this, we propose a hierarchical MA-LLM framework that modularly separates reasoning, perception, and execution. This hierarchical structure facilitates flexible task allocation, semantic-aware motion planning, and robust manipulation within dynamic, heterogeneous robotic systems.

\begin{figure*}[ht]
    \centering
    \includegraphics[width=0.8\linewidth]{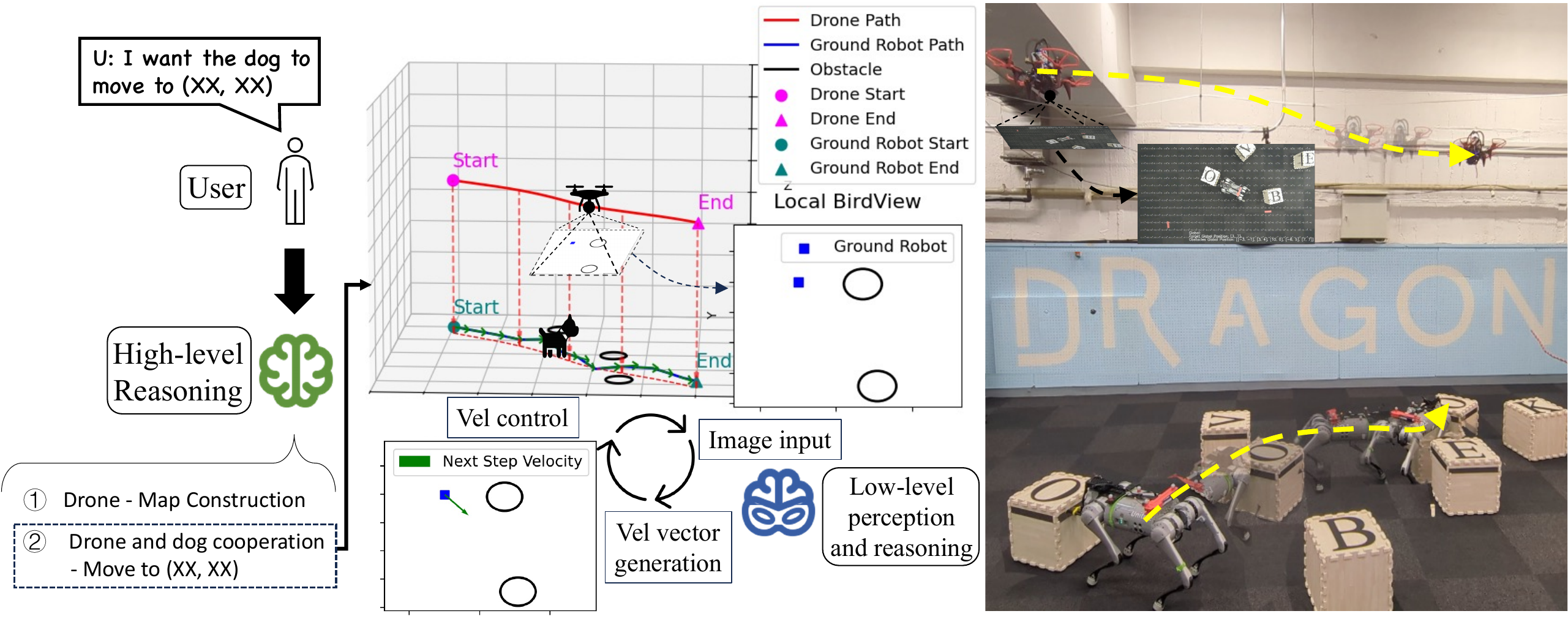}
    \captionsetup{skip=0pt}
    \caption{Overview of the proposed hierarchical language model framework integrated into an aerial-ground robotic system. In the sub-task ``move to (XX, XX)", the aerial robot follows an optimized global path while continuously capturing bird-view images. These images are processed into semantic information that guides the ground robot’s real-time local navigation, implicitly allowing the ground robot to follow the aerial robot’s position.}
    \label{fig:archi}
    \vspace{-4mm}
\end{figure*}

In our proposed hierarchical MA-LLM framework (Figure~\ref{fig:archi}), task execution is structured into three functional layers: a reasoning layer that decomposes user commands into task flows, a perceptual layer that extracts semantic information from aerial images, and an execution layer that performs motion through aerial-ground collaboration.
The execution process is guided by a leader-follower mechanism, where the aerial robot plans an optimized global path and provides a continuous stream of bird-view images. The ground robot uses the center of these images—corresponding to the aerial robot’s projected ground position—as a dynamic navigation anchor, allowing itself to follow the aerial path even when the target object is temporarily out of view.

\textcolor{black}{The main contributions of this paper are as follows: 
\begin{enumerate}
    \item We propose a hierarchical multimodal MA-LLM framework for heterogeneous aerial–ground robot systems, which tightly couples LLM-based task reasoning with VLM-based semantic perception. The aerial robot provides global semantic guidance from bird-view images, while the ground robot executes local navigation and manipulation. This bridges high-level reasoning and low-level execution, enabling scalable and adaptive task performance.
    \item We introduce a GridMask-based fine-tuning method for VLMs that outperforms conventional coordinate-supervised baselines by a large margin, reducing localization errors by 78\% on the same-scale training set. This enables accurate semantic labeling and object localization from bird-view images without depth sensing, thereby substantially supporting the construction of the global semantic map and ensuring reliable fine-grained manipulation.
    \item We conduct extensive evaluations in both simulation and real-world settings, demonstrating (i) zero-shot generalization to unseen objects, semantic configurations, and task compositions, (ii) empirical comparisons with representative Deep Reinforcement Learning (DRL) multi-robot baselines on a target-search subtask, and (iii) robust long-horizon aerial–ground collaboration validated through real-world experiments.
\end{enumerate}}

\section{Related Works}
\subsection{Heterogeneous Multi-Robot Systems}
HMRS leverages diverse agent capabilities to address complex tasks. One of the earliest works in this area is Parker's ALLIANCE architecture \cite{firsthmrs}, which introduced fault-tolerant behavior-based coordination. Follow-up studies examined distributed task allocation and coordination \cite{iocchi2003distributed,MRTA,notomista2019optimal}. \textcolor{black}{More recently, deep Multi-Agent Reinforcement Learning (MARL) has been explored to improve scalability and adaptability in decentralized settings \cite{hetero_multiagent_tro, MARLreview, MARL, chen2025target}}, particularly for navigation and search tasks.

Beyond these approaches, mapping frameworks in HMRS (e.g., shared-SLAM \cite{sharedslam,lamp2}) have enabled teams of robots to maintain consistent global spatial representations. However, such systems mainly facilitate spatial alignment and navigation, still rely on manually defined rules or domain knowledge \cite{semanticmapreview,semanticmapreview2}. Although such methods can address navigation and search at the local policy level, they remain limited in semantic reasoning and compositional planning, whereas our framework explicitly leverages LLMs and VLMs to bridge global semantic reasoning with low-level execution.

\subsection{\textcolor{black}{Task Planning with Large Language Models}}
\textcolor{black}{Table~\ref{tab:comparisonplanning} summarizes the key trade-offs among representative frameworks for task reasoning and allocation.} Traditional state machine-based approaches such as Finite State Machines (FSMs) and Behavior Trees (BT) \cite{bt} offer minimal engineering overhead and high interpretability, 
but their rigid structures severely limit generalization to novel or long-horizon tasks. 
Classical symbolic planners such as STRIPS \cite{strips} and PDDL \cite{pddl} provide greater task flexibility and strong explainability, 
yet they require manual modeling of domain knowledge and action schemas, leading to moderate engineering costs and limited scalability in dynamic or open-ended scenarios. 

In contrast, LLMs offer a new paradigm for task planning by directly leveraging natural language as the interface between user intent and robot actions. 
LLM-based frameworks such as SayCan \cite{saycan}, ProgPrompt \cite{progprompt}, and Tidybot \cite{Wu_2023} have shown that high-level user instructions can be decomposed into executable skills in constrained settings. 
More recent works, including COHERENT \cite{coherent}, have demonstrated that LLMs can also facilitate task allocation and coordination in multi-robot systems. 

Building on these advances, our method integrates LLMs not only for hierarchical task decomposition but also for global semantic map construction by aggregating aerial observations, thereby enabling long-horizon reasoning and heterogeneous aerial–ground collaboration in dynamic environments.

\begin{table*}[htbp]
\centering
\caption{\textcolor{black}{Comparison of reasoning and allocation frameworks for multi-robot task decomposition.}}
\begin{tabular}{lccc}
\toprule
\textbf{Method} & \textbf{\makecell{Engineering Overhead}} & \textbf{\makecell{Task Generalization}} & \textbf{\makecell{Explainability}} \\
\midrule
FSM/BT \cite{bt} & Low & Low & High \\
STRIPS/PDDL \cite{strips,pddl} & Moderate & Moderate & High \\
\textbf{LLM-based} & Low & High & Moderate-High \\
\bottomrule
\label{tab:comparisonplanning}
\end{tabular}
\end{table*}

\begin{figure*}[ht]
    \centering
    \includegraphics[width=0.88\textwidth]{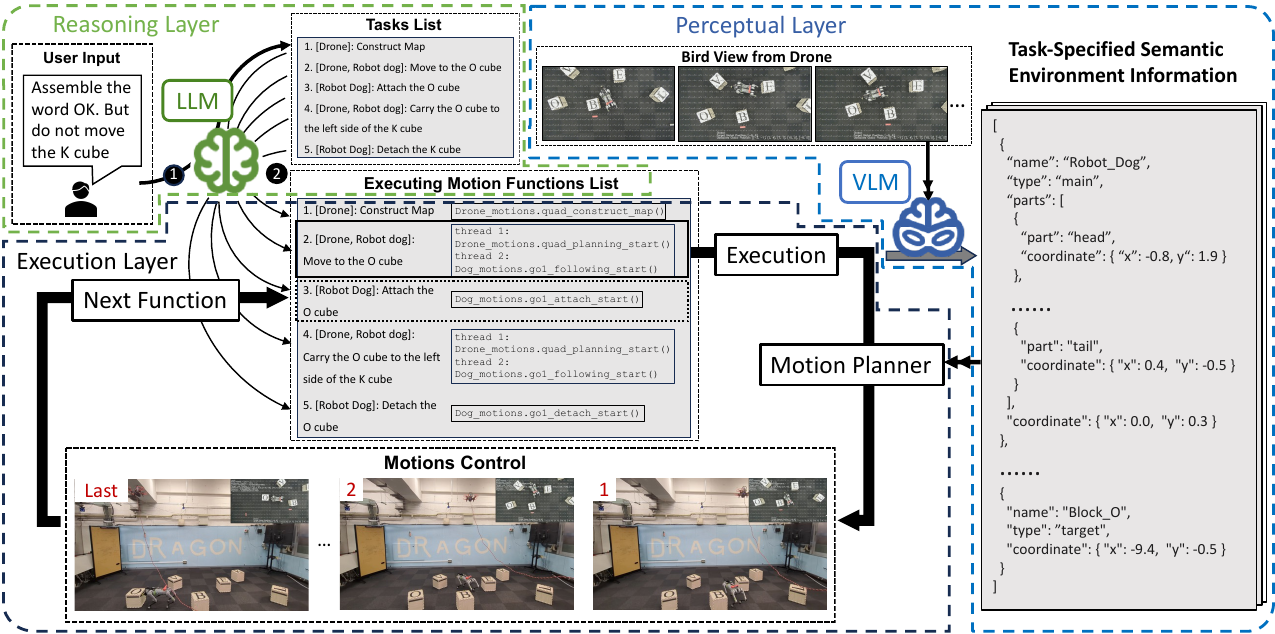}
    \captionsetup{skip=0pt}
    \caption{An overview of the workflow for a long-horizon task using the hierarchical MA-LLM framework. The task starts with the instruction: ``Assemble the word OK, but do not move K.'' 1) The LLM decomposes the command and maps sub-tasks to motion functions for the aerial and ground robots. 2) The aerial robot visits multiple viewpoints to collect local maps, which the LLM integrates into a global semantic map. 3) Once the map is ready, both robots coordinate to reach the ``O'' cube. 4) The aerial robot follows a task-specific global path, while the VLM processes GridMask-enhanced bird-view images. 5) The ground robot uses this semantic input from the VLM to complete its assigned sub-task via local planning.}
    \vspace{-4mm}
    \label{fig:multimodal}
\end{figure*}

\subsection{Vision-Language Models for Robotic Perception}

Vision has long served as the primary modality for robotic perception, enabling scene understanding through object detection, recognition, and spatial reasoning. 
\textcolor{black}{Progress in vision-based learning has further enhanced robots’ perception, particularly for aerial platforms~\cite{xiao2025vision}, pushing their agility and autonomy beyond classical pipelines, but challenges remain in semantic generalization.}

Recent advances in VLMs, such as CLIP~\cite{clip} and Vision Transformers~\cite{visontransformer}, 
bridge visual inputs with semantic concepts, providing a richer understanding of task-relevant entities. 
However, these general-purpose VLMs still lack precise spatial reasoning, which is essential for robotic navigation and manipulation. 
In parallel, specialized Vision-Language-Action (VLA) models like the RT-series~\cite{rt2,rtx} integrate perception with action grounding for end-to-end robot control, 
but they continue to face challenges in fine-grained localization.

To address these limitations, we employ GridMask-based fine-tuning, which strengthens the VLM’s ability in spatially aware recognition and semantic labeling. 
This fine-tuned VLM serves as the semantic engine of our aerial–ground framework, supporting both global semantic mapping and local navigation and manipulation tasks.

\section{Methods}

To enable reliable task execution in our aerial-ground robotic system, we adopt a hierarchical MA-LLM framework. As illustrated in Figure~\ref{fig:multimodal}, the system is organized into three functional layers: a reasoning layer for task decomposition and allocation, and global map construction, a perceptual layer that extracts semantic understanding from aerial imagery, and an execution layer that performs robot actions through motion functions grounded in the reasoning and perception outputs.

The framework consists of the following components:
\begin{itemize}
    \item \textit{LLM-based reasoning layer}: Interprets natural language commands, decomposes them into structured sub-tasks, and selects appropriate motion functions for each robot. It also constructs and updates a global semantic map by integrating spatial-contextual information gathered during execution.

    \item \textit{VLM-based perceptual layer}: Analyzes GridMask-enhanced aerial images to extract object-level semantics, including positions, orientations, and categories. These semantic cues provide environmental context for both initial mapping and real-time navigation support.

    \item \textit{Execution layer}: Executes motion functions selected by the reasoning layer, each corresponding to a decomposed sub-task. These functions dynamically interact with the perceptual layer to obtain task-specific semantic information. For example, in global map construction, this layer gathers local semantic maps for global map integration by the reasoning layer; during local navigation and manipulation, local semantic information is extracted in real-time to guide ground motions. This layer acts as the bridge between high-level plans and grounded actions, coordinating aerial-ground behaviors across perception and reasoning.

\end{itemize}

\subsection{LLM-based Reasoning Layer Design}

To enhance the reasoning capability of the LLM-based layer, we design task-specific prompts tailored to its functional responsibilities within the aerial-ground robotic system. The summarized prompts are as follows:
\subsubsection{Task Decomposition and Allocation}
To enable reasoning across heterogeneous agents, we adopt a role-specific prompt design. The LLM is instructed to act as a task decomposer and allocator, mapping high-level goals to robot-specific sub-tasks based on their capabilities.
\begin{codebox}
\textcolor{codegreen}{\footnotesize \texttt{\# Duty clarify}}\\
\colorbox{codeblue}{%
    \begin{minipage}{\dimexpr\textwidth}
        \footnotesize \texttt{You are a task decomposer and allocator for a heterogeneous multi-robot system. Decompose complex tasks into sub-tasks for each robot according to their abilities.}
    \end{minipage}%
}\\
\textcolor{codegreen}{\footnotesize \texttt{\# Robots' abilities}}\\
\colorbox{codeblue}{%
    \begin{minipage}{\dimexpr\textwidth}
    \setlength{\parindent}{1em}
        \footnotesize{\texttt{\hspace{-1.3em}- Drone:\\
        \indent 1. Construct the map. (Must be the first step before other sub-tasks)\\
        - Robot Dog:\\
        \indent 1. Attach/Detach to an object.\\
        - Drone and Robot Dog Cooperation:\\
        \indent 1. Move to a specified location.\\
        \indent 2. Carry an object to a specified location.}}
            
    \end{minipage}%
}
\end{codebox}

\subsubsection{Motion Functions Mapping}
 To ensure the LLM generates executable action plans, we provide the available motion functions for each robot via the prompt. This prevents invalid actions and constrains the reasoning process to grounded, system-supported functions.

\begin{codebox}
\textcolor{codegreen}{\footnotesize \texttt{\# Duty clarify}}\\
\colorbox{codeblue}{%
    \begin{minipage}{\dimexpr\textwidth}
    \setlength{\parindent}{1em}
    \footnotesize\texttt{\hspace{-1.3em}You are responsible for choosing the motion function to finish the task.}
    \end{minipage}%
}
\\
\textcolor{codegreen}{\footnotesize \texttt{\# Robots' motion functions}}\\
\colorbox{codeblue}{%
    \begin{minipage}{\dimexpr\textwidth}
        \setlength{\parindent}{1em}
        \footnotesize \texttt{\hspace{-1.3em}- Drone:\\
            \indent 1. quad\_construct\_map()\\
            - Robot Dog:\\
            \indent 1. go1\_attach\_start("name\_of\_target\_object")\\
            \indent 2. go1\_detach\_start("name\_of\_target\_object")\\
            - Drone and Robot Dog Cooperation:\\
            \indent Thread 1: drone.quad\_planning\_start("task")\\
            \indent Thread 2: dog.go1\_following\_start("task")
            }   
    \end{minipage}%
}
\end{codebox}

\subsubsection{Global Semantic Map Integration}
These chain-of-thought prompts are used to improve the model's ability to analyze local maps and construct a global map.
\begin{codebox}
\textcolor{codegreen}{\footnotesize \texttt{\# Duty clarify}}\\
\colorbox{codeblue}{%
    \begin{minipage}{\dimexpr\textwidth}
    \setlength{\parindent}{1em}
    \footnotesize\texttt{\hspace{-1.3em}You need to construct a global map by logically integrating all provided local map data.}
    \end{minipage}%
}
\\
\textcolor{codegreen}{\footnotesize \texttt{\# Chain-of-thought}}\\
\colorbox{codeblue}{%
    \begin{minipage}{\dimexpr\textwidth}
    \setlength{\parindent}{1em}
    \footnotesize{\texttt{\hspace{-1.3em}- Cross-map Validation:\\
            \indent 1. Always cross-check objects across multiple local maps; do not rely on a single observation.\\
            - Error Identification:\\
            \indent 1. Detect and exclude erroneous entries.\\
            \indent 2. If an object is only present in a single local map and absent from others, where logically it should be visible, remove it.\\
            - Conflict Resolution:\\
            \indent 1. If objects appear unrealistically close, identify the mislabeled object and correct or remove it.\\
            \indent 2. Resolve conflicting labels by majority voting across maps or flags.}}
    \end{minipage}%
}
\end{codebox}

This layer serves as the top-level reasoning layer, translating user commands into modular functions for downstream execution. Notably, although the prompts and APIs described above are tailored for our aerial-ground robotic platform, the proposed LLM-based reasoning framework is inherently modular and extensible. The division of labor (task decomposition, ability declaration, and function mapping) is agent-agnostic and can be adapted to a wide range of heterogeneous robot teams. For new robot types or novel tasks, only the corresponding capability and action definitions need to be supplied in the prompt, while the core reasoning and allocation logic remain unchanged. This design facilitates transferability across robot platforms and supports rapid extension to unseen tasks, provided that the required actions are defined in the prompt or API library, in line with recent advances in modular prompt engineering and prompt libraries for robotic task generalization~\cite{codeaspolicy, enhancingllmdmp, promptbook}.

\subsection{VLM-based Perceptual Layer Design}

To enhance the perceptual capabilities of the VLM-based layer, we employ prompt engineering to improve visual–language alignment and structured output \textcolor{black}{\cite{white2023prompt}}, and \textcolor{black}{fine-tune the model on a GridMask-augmented 2D recognition dataset to achieve precise object detection.}

First, we design role-specific prompts to guide the VLM in outputting structured environmental information clearly and accurately:

\begin{codebox}
\textcolor{codegreen}{\footnotesize \texttt{\# Duty Clarification}}\\
\colorbox{codeblue}{%
    \begin{minipage}{\dimexpr\textwidth}
    \setlength{\parindent}{1em}
        \footnotesize\texttt{\hspace{-1.3em}You are an environment describer and classifier. Your task is to identify all objects in the environment, provide their names and coordinates in a structured JSON format, and classify them according to the provided task requirements.}
    \end{minipage}%
}\\
\textcolor{codegreen}{\footnotesize \texttt{\# Detailed Instructions}}\\
\colorbox{codeblue}{%
    \begin{minipage}{\dimexpr\textwidth}
    \setlength{\parindent}{1em}
        \footnotesize\texttt{\hspace{-1.3em}- The environment is overlaid with a dense coordinate grid, with the origin at the center. The x-axis extends horizontally from left to right, and the y-axis extends vertically from bottom to top. The coordinate (0, 0) represents the central reference point.\\
        - Depending on the given task, classify each object by labeling it into one of four categories: main, target, landmark, or obstacle.\\
        - Important Note:\\
        \indent 1. For local navigation tasks, do not use global coordinates directly. Instead, if the task involves moving to a specified global coordinate (XX, XX), set this target as (0, 0) to indicate that the ground robot should follow relative movements guided by the aerial robot.\\
        \indent 2. For map construction, you do not need to make the labeling classification.}
    \end{minipage}%
}
\end{codebox}

While prompts guide the VLM's response format and labeling logic, spatial localization remains a challenge. To address the challenge of accurate spatial localization, we introduce GridMask—a structured visual cue that overlays a dense coordinate grid on bird’s-eye-view images captured by the aerial robot. Each grid cell measures $s \times s$ pixels (typically $s = 80$), offering a balance between spatial resolution and semantic clarity.

Formally, for an input image of resolution $w \times h$, we define the grid in a Cartesian coordinate system centered at the image midpoint, with the $x$-axis pointing rightward and the $y$-axis pointing upward. Each grid cell measures $s \times s$ pixels, and the pixel coordinates of each grid vertex $(x_i, y_j)$ are given by
\begin{equation}
x_i = -\frac{w}{2} + i s + \alpha,\qquad
y_j = \frac{h}{2} - j s - \beta,
\end{equation}
\textcolor{black}{where $i,j \in \mathbb{Z}$ are integer indices such that 
$0 \leq i < \lfloor w/s \rfloor$, $0 \leq j < \lfloor h/s \rfloor$, 
and $\alpha,\beta \in \mathbb{R}$ are user-defined offsets that shift the grid uniformly to ensure symmetry around the origin.}

To ensure generalization across diverse imaging configurations, the grid density and scaling are dynamically determined by the camera resolution, field of view (FOV), and flight altitude. The number of grid divisions is
\begin{equation}
N_{\text{grid}} = \frac{w}{s},
\end{equation}
where $w$ is the image width in pixels.

The physical ground width covered by the image is
\begin{equation}
W_{\text{real}} = 2 h_{\text{cam}} \cdot \tan \left( \frac{\text{FOV}}{2} \right),
\end{equation}
where $h_{\text{cam}}$ is the camera altitude and $\text{FOV}$ is the horizontal field of view in radians.

The real-world size corresponding to each grid cell is then defined as
\textcolor{black}{\begin{equation}
\kappa = \frac{W_{\text{real}}}{N_{\text{grid}}}.
\label{eq:prop}
\end{equation}}

\begin{figure}[t]
    \centering
    \includegraphics[width=1\linewidth]{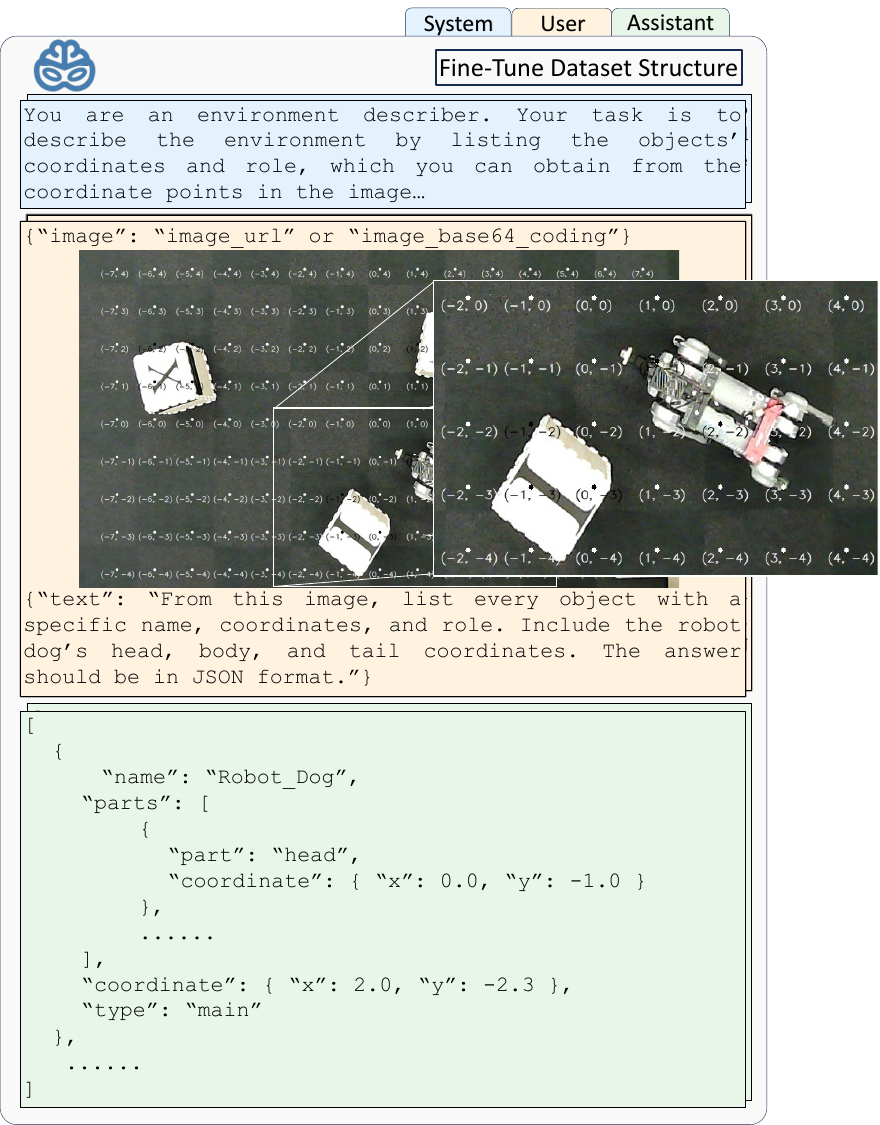}
    \captionsetup{skip=0pt}
    \caption{\textcolor{black}{An illustration of the fine-tuning dataset, showing the system prompt, user instruction with GridMask-based bird-view image input, and the ideal output in structured JSON format.}}
    \label{fig:fine-tune}
\end{figure}

As shown in Figure~\ref{fig:fine-tune}, this principled \textcolor{black}{dataset structure} enables the VLM to associate visual features with absolute 2D coordinates in a data-driven manner, while ensuring that the spatial cues are resolution-agnostic and readily adaptable to diverse camera settings.

Based on this structured dataset, we fine-tune the VLM using paired textual and visual information together with corresponding object coordinates. After fine-tuning, the VLM is able to robustly identify objects’ spatial positions and roles in a zero-shot manner, classified as follows:
\begin{itemize}
    \item \texttt{main}: The primary agent performing the task, typically the ground robot. Objects carried by the robot are also labeled \texttt{main}.
    \item \texttt{target} \& \texttt{landmark}: \texttt{target} denotes the direct goal position; \texttt{landmark} serves as a spatial reference, further classified into four directional indicators (\texttt{front}, \texttt{back}, \texttt{left}, and \texttt{right}).
    \item \texttt{obstacle}: This field represents all objects that obstruct motion paths. During task execution, \texttt{landmark} simultaneously serves as \texttt{obstacle}.
\end{itemize}

\textcolor{black}{As the perceptual layer, it provides structured and precise semantic observations that serve as the foundation for planning at all levels.}

\subsection{Execution Layer Design - Motion Functions Pre-Programming}
Acting as the executing layer, it bridges the reasoning and perceptual layers, ensuring the execution of low-level motion functions.
To realize this execution, we define a set of pre-programmed motion functions that enable task-level behaviors in the aerial-ground robotic system, as described below.

\subsubsection{Aerial Robot - Global Semantic Map Construction}
\begin{figure*}[htbp]
    \centering
    \includegraphics[width=\linewidth]{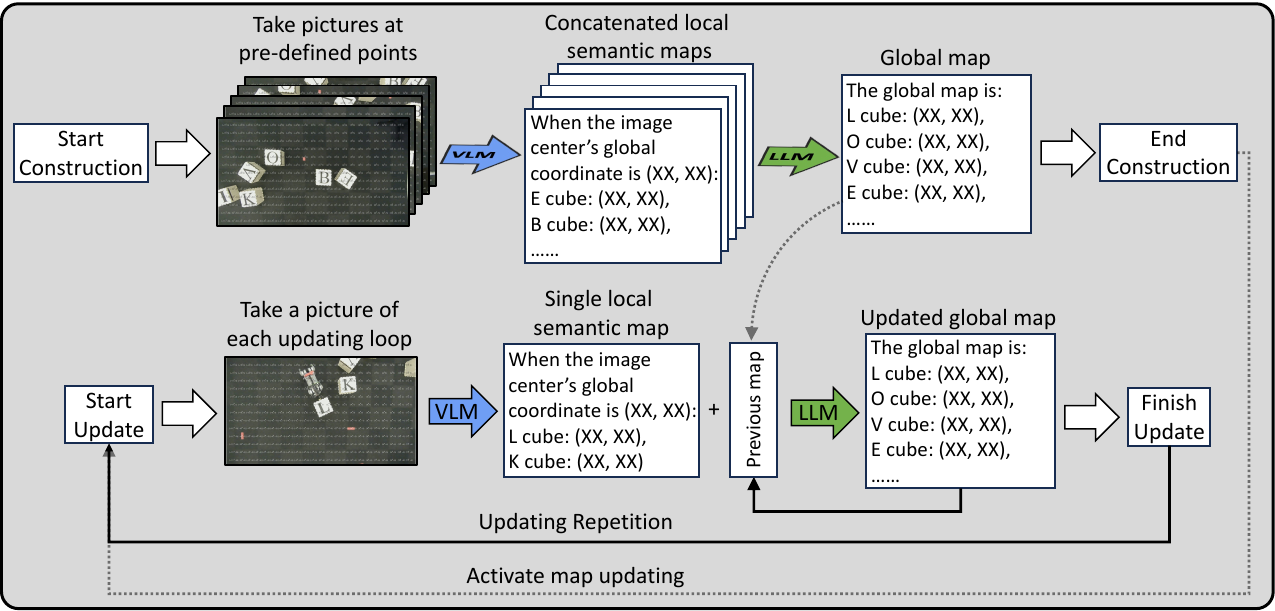}
    \captionsetup{skip=0pt}
    \caption{\textcolor{black}{A workflow of the \texttt{quad\_construct\_map()} function. Local semantic maps derived from aerial images are integrated by the LLM to form a semantic map, which is subsequently updated with 15s interval during task execution.}}
    \label{fig:map}
\end{figure*}
The \texttt{quad\_construct\_map()} function enables high-level spatial understanding by fusing perception, motion context, and language reasoning. It serves as the semantic foundation for downstream planning by generating a global map from aerial image observations. The process consists of the following steps and is illustrated in Figure \ref{fig:map}:

\begin{itemize}
    \item Local Map Generation: The aerial robot executes a predefined waypoint path to collect bird-view images. Each image is processed by the VLM to produce local semantic maps containing object classes and positions.
    \item Global Map Integration: The LLM receives a concatenated textual representation of all local semantic maps and infers a spatially consistent global map. This process involves cross-map validation, confidence reasoning, and label disambiguation. Meanwhile, the global map will keep updating during task execution.
\end{itemize}

\subsubsection{Aerial Robot - Global Path Planning}
\label{subsec:globalplan}

As the leader in the aerial-ground robotic system, the aerial robot plans a globally optimal path that not only avoids obstacles but also dynamically guides the ground robot. The function \texttt{quad\_planning\_start()} (the aerial robot’s global path planner) enables autonomous navigation and obstacle avoidance by generating a collision-free path from a global semantic map. This path guides the ground robot by defining a dynamically updated spatial target (zero point), allowing it to follow the aerial robot through task-relevant areas with global awareness.

Given the semantic map from the LLM and object classification from the VLM (\texttt{type:  target, main, obstacle, landmark}) with their coordinates (\texttt{coordinate}), the system constructs a task-specific cost function for global path optimization. This path ensures smoothness and collision avoidance through spline interpolation and optimization techniques. Details of this optimization are described mathematically below.

\textcolor{black}{At the beginning, the initial control points $\mathbf{C} = \{ \mathbf{c}_0, \mathbf{c}_1, \ldots, \mathbf{c}_n \}$ 
are defined from semantic anchors provided by the VLM: 
the start $\mathbf{c}_0$ is aligned with the robot’s current coordinate \texttt{main}, 
the end $\mathbf{c}_n$ corresponds to the \texttt{target} (or designated \texttt{landmark}) coordinate.
\texttt{obstacle} coordinates $\mathbf{O} =\{ \mathbf{o}_0, \mathbf{o}_1, \ldots, \mathbf{o}_{m-1} \}$ are not directly used as control points but instead define repulsive regions 
that contribute to the collision cost in the global objective.}

\textcolor{black}{Subsequently}, the optimized path, represented by the sequence of control points $\mathbf{C} = \{ \mathbf{c}_0, \mathbf{c}_1, \ldots, \mathbf{c}_n \}$, is expressed as a continuous curve through the B-spline.
\begin{equation}
    \mathbf{S}(u; \mathbf{C}) = \sum_{i=0}^{n} N_{i,k}(u) \mathbf{c}_i,
\end{equation}
\textcolor{black}{where $N_{i,k}(u)$ are the B-spline basis functions of order $k$, and $u \in [0,1]$ serves as the parametric variable along which the curve is sampled.}

\textcolor{black}{Then}, the optimization is formulated as minimizing a cost function $J_{\text{global}}$ over the control points $\mathbf{C}$ \textcolor{black}{and \texttt{obstacles} $\mathbf{O}$}: 
\begin{equation} 
\mathbf{C}^* = \arg\min_{\mathbf{C}} J_{\text{global}}(\mathbf{S}(u; \mathbf{C}), \mathbf{O}), 
\end{equation} 
where the cost function $J_{\text{global}}$ balances path length, smoothness, and obstacle avoidance. Its detailed formulation is provided in Appendix~\ref{app:globalpath}.

\textcolor{black}{In the last step}, the real-world waypoints $\mathbf{S}_{\text{real}}$ are obtained by sampling this optimized B-spline and projecting them into real-world coordinates using the spatial scaling factor $s_{\text{cell}}$ from Equation~(\ref{eq:prop}):
\begin{equation}
    \mathbf{S}_{\text{real}}(u) = \kappa \cdot \mathbf{S}(u; \mathbf{C}^*).
\end{equation}
These waypoints form the basis for semantic guidance, with the aerial robot’s ground-projected position serving as a continuously updated local navigation anchor for the ground robot.

\subsubsection{Ground Robot – Local Semantic Navigation and Manipulation}
\label{subsec:localplan}

Inspired by the classical Dynamic Window Approach (DWA) \cite{fox1997dynamic}, the ground robot acts as the follower in the aerial–ground system and performs iterative local navigation \textcolor{black}{within a dynamic workspace defined by the aerial robot’s projected path and semantic cues}, implemented through \texttt{go1\_following\_start()}.

\textcolor{black}{Similar to the global path planning}, the semantic input for this local planning is derived from the VLM, which processes bird-view images to generate a task-specific local semantic map centered at the aerial robot’s projection point. The map includes object-level annotations (\texttt{type: target, main, obstacle, landmark}) and part-level annotations for the ground robot (\texttt{parts: head, body, tail}) with their coordinates (\texttt{coordinate}). These annotations enable the local motion planner to estimate the robot's current pose, obstacle distribution, and target-relative position within the local frame. 

\textcolor{black}{Initially, the semantic labels directly shape the local cost function: 
\texttt{main} coordinate $\mathbf{M}$ and \texttt{targets} (or \texttt{landmarks}) coordinate $\mathbf{T}$ define attractive directions, 
\texttt{obstacle} coordinates $\mathbf{O} =\{ \mathbf{o}_0, \mathbf{o}_1, \ldots, \mathbf{o}_{m-1} \}$ contribute to repulsive penalties, 
and the robot’s body-part annotations (\texttt{head}, \texttt{body}, \texttt{tail}) 
provide current orientation to maintain feasible kinematics.} 

Accordingly, the local motion planner generates candidate velocity directions $\theta \in \{\theta_0, \theta_1, \ldots, \theta_{n-1}\}$ uniformly distributed over $360^\circ$, 
and selects the optimal direction $\theta^*$ by minimizing the task-specific cost $J_{\text{local}}$:

\textcolor{black}{\begin{equation}
\theta^* = \arg\min_{\theta} J_{\text{local}}(\theta, \mathbf{M},\mathbf{T},\mathbf{O}).
\end{equation}}

where the cost function $J_{\text{local}}$ jointly balances target deviation, central alignment, obstacle avoidance, and window restriction. Its detailed formulation is provided in Appendix~\ref{app:localdirection}. 

Once the optimal direction $\theta^*$ is selected, the robot adjusts its orientation to it and moves forward or backward based on the relative distance to the target in its local coordinate frame, calculated from semantic annotations and scaled by $\kappa$ as defined in Equation~(\ref{eq:prop}). 
The iterative local planner terminates when both the positional (distance to target below a predefined threshold) and orientation alignment (within a certain angular tolerance) criteria are met.

Through this mechanism, the ground robot adaptively navigates toward the target while remaining synchronized with the aerial robot, achieving cooperative semantic navigation.

\subsubsection{Ground Robot - Attach and Detach}
The ground robot is equipped with a servo-driven magnetic attachment device. This device allows the robot to attach to or detach from objects located in front of it, enabling interactions between blocks or other objects with magnetism.

By using function \texttt{go1\_attach\_start()} and \texttt{go1\_detach\_start()}, the system can control the servo to actuate the ground robot's magnet switch, toggling between attachment and detachment during execution.

\subsubsection{Aerial Robot and Ground Robot - Engineering Supplement}
\label{subsec：engineering}
In addition to the core recognition, reasoning, and planning mechanisms, several practical engineering enhancements were implemented to ensure the robust operation of the aerial-ground robotic system.
\begin{itemize}
    \item Aerial Robot - Adaptive Trajectory Execution: During trajectory execution, the aerial robot dynamically adjusts its movement based on real-time feedback from the VLM. Specifically, it predicts and monitors the ground robot's position and actively waits as necessary, maintaining the ground robot's position near the center of its camera view. This adaptive strategy ensures continuous visual coverage and stable guidance.
    \item \textcolor{black}{Ground Robot - Rotation Stability}: When the ground robot performs in-place rotations, the rotation direction (clockwise or counterclockwise) is dynamically determined based on semantic spatial information from the VLM. This adaptive decision-making minimizes the risk of collision with surrounding obstacles during orientation adjustments.
    \item \textcolor{black}{Ground Robot - Carrying Stability}: During object transportation tasks, the ground robot continuously utilizes VLM feedback to monitor whether the carried object remains properly attached. If the VLM detects that the object is no longer in the desired carrying state, the task execution initiates an immediate rollback procedure to reattempt the previous attaching sub-task and ensure successful task completion.

\end{itemize}

These practical engineering solutions complement the high-level semantic reasoning and planning layers, significantly enhancing the overall robustness and reliability of the robotic system during real-world operations.

\section{Experiments and Discussions}
To comprehensively assess the proposed framework, we conduct experiments and discussions along three key axes:

\begin{itemize}
    \item Evaluation of semantic perception improvements via GridMask-based VLM fine-tuning.
    \item \textcolor{black}{Test zero-shot transferability and generalizability of the system to different environments and objects in simulation, also discuss the low-level target search ability of the system by comparing with other baseline methods.}
    \item Assessment of real-world task performance in an aerial-ground heterogeneous robotic system.
\end{itemize}

\subsection{\textcolor{black}{System and Environment Implementation}}
\label{subsec:system-env}

In our experiments, the system is deployed on a heterogeneous aerial–ground platform \cite{yunong}, comprising a Unitree Go1 quadruped and a custom quadrotor equipped with onboard SLAM and a nadir-facing camera for bird-view perception. Details are provided in Table~\ref{tab:system_specs}.

\begin{table*}[htbp]
\centering
\caption{System specifications of the heterogeneous aerial–ground robotic platform. }
\label{tab:system_specs}
\begin{tabular}{lcc}
\toprule
\textbf{Component} & \textbf{Specification} & \textbf{Function} \\
\midrule
Aerial Robot & Custom quadrotor, SLAM-capable & Capture bird-view images \\
Ground Robot & Unitree Go1 with magnetic gripper & Perform ground locomotion and block manipulation \\
Onboard PC & Vim4 (Linux) $\times$ 2, 8GB RAM & Run ROS nodes and onboard SLAM/control modules \\
Central Host & Notebook PC (Linux) & Run API calls, planners, and dispatch robot commands \\
Camera & ELP USB wide-angle, FOV $90^\circ$, 1080p & Provide visual input to the VLM \\
Computation & Google Cloud API (LLM/VLM inference) & Execute task reasoning and semantic perception \\
Communication & Wireless ROS communication & Enable aerial–ground coordination \\
\bottomrule
\end{tabular}
\end{table*}

The LLM (reasoning) and VLM (perception) are accessed through the Google Cloud API from a Linux host over a wireless network. A high-level overview of deployment choices and measured request latencies is summarized in Table~\ref{tab:llm_vlm_params}.

\begin{table*}[htbp]
\centering
\caption{Deployment settings and runtime latency of LLM and VLM modules. 
The LLM uses moderate temperature to preserve reasoning diversity, while the VLM adopts deterministic decoding for stable perception.}
\label{tab:llm_vlm_params}
\begin{tabular}{lccc}
\toprule
\textbf{Model} & \textbf{Key Parameters} & \textbf{Deployment} & \textbf{Representative Latency (s)} \\
\midrule
\makecell[l]{LLM (Reasoning)\\ \texttt{Gemini-2.0-pro}} 
& \makecell[l]{Max tokens: 8192 \\ Temperature: 0.5} 
& Google Cloud API 
& \makecell[l]{Task planning: $\sim$4.0 \\ Semantic map: $\sim$15.0 \\ Motion function selection: $\sim$1.5} \\
\midrule
\makecell[l]{VLM (Perception)\\ \texttt{Gemini-2.0-flash}} 
& \makecell[l]{Max tokens: 8192 \\ Temperature: 0.0} 
& Google Cloud API 
& \makecell[l]{Semantic labeling and localization: $\sim$3.0} \\
\bottomrule
\end{tabular}
\end{table*}

The environment configurations for simulation and real-world experiments are summarized in Table~\ref{tab:env_setup}.

\begin{table*}[htbp]
\centering
\caption{Configuration of simulation and real-world experimental environments. 
Simulation is conducted in Gazebo, and real-world trials are performed in a laboratory indoor environment.}
\label{tab:env_setup}
\begin{tabular}{lcc}
\toprule
\textbf{Environment} & \textbf{Specification} & \textbf{Details} \\
\midrule
Simulation & Gazebo-based 3D environment & $5 \times 5 \times 3$ m open workspace (no enclosing walls) \\
Real-world & Laboratory indoor environment & $5 \times 5 \times 3$ m workspace \\
Drone Altitude & Constant flight height & 2 m above ground \\
Task Objects & Color-coded and letter-labeled blocks & Cube edge length: $0.3$ m \\
\bottomrule
\end{tabular}
\end{table*}

\subsection{Experiment for the GridMask-enhanced VLM-based Perception}

A key objective of this experiment is to benchmark our GridMask-based fine-tuning approach against coordinate-supervision baselines for VLM-driven spatial localization, particularly in the regime of low-rank and small-sample fine-tuning \cite{lora}, where label efficiency and precision are critical for robotics deployment. We fine-tune the \texttt{Gemini-2.0-flash} model, with details of its architecture and merits available in the official paper \cite{geminiteam2024geminifamilyhighlycapable}.

\textcolor{black}{We compare two principal categories:  
(i) {Coordinate-only supervision}, where the VLM is fine-tuned to regress object coordinates directly from images (standard baseline for open-vocabulary localization \cite{clip,owl}); and  
(ii) {GridMask-augmented supervision}, where dense spatial overlays are added during both fine-tuning and inference, explicitly encoding positional cues.  
The experimental groups and their correspondences are summarized in Table~\ref{tab:baselines-mapping}.}

\begin{table*}[htbp]
    \centering
    \caption{\textcolor{black}{Mapping of experimental groups to baseline categories. 
    GridMask-augmented datasets share the same text queries and reference annotations as the original datasets, but with images overlaid by GridMask. 
    Dataset structure is illustrated in Figure~\ref{fig:fine-tune}.}}
    \begin{tabular}{p{6cm}p{8cm}p{2cm}}
        \toprule
        \hspace{2mm}\textbf{Model} & \textbf{\textcolor{black}{Fine-tuning Dataset and Evaluation Setting}} & \textbf{Type} \\
        \midrule
        \hspace{2mm}\textbf{400G-FT $|$ GridMask} & \makecell[l]{Fine-tune on 400 GridMask-augmented images
\\Eval on GridMask images} & {Ablation} \\
        \midrule
        \hspace{2mm}\textbf{200G-FT $|$ GridMask} & \makecell[l]{Fine-tune on 200 GridMask-augmented images\\Eval on GridMask images} & \makecell[l]{GridMask-\\Augmented} \\
        \midrule
        \hspace{2mm}200G-FT $|$ NoGridMask & \makecell[l]{Fine-tune on 200 GridMask-augmented images\\Eval on original images (No GridMask)} & Ablation \\
        \midrule
        \hspace{2mm}200-FT $|$ GridMask & \makecell[l]{Fine-tune on 200 original images\\Eval on GridMask images} & Ablation \\
        \midrule
        \hspace{2mm}200-FT $|$ NoGridMask & \makecell[l]{Fine-tune on 200 original images\\Eval on original images (No GridMask)} & \makecell[l]{Coordinate\\Baseline} \\
        \midrule
        \hspace{2mm}Base $|$ GridMask / NoGridMask & No task-specific fine-tuning, direct eval & Reference \\
        \bottomrule
    \end{tabular}
    \label{tab:baselines-mapping}
\end{table*}

Besides coordinate supervision and GridMask-based approaches which were evaluated in our experiments, other mainstream VLM grounding strategies include: (i) region-based annotation (such as bounding boxes or segmentation masks) is widely used to provide explicit spatial cues and has achieved strong results in open-vocabulary detection and localization \cite{regionclip, zareian2021open}. However, in our tabletop scenario, current mass point coordinate supervision is functionally equivalent to bounding box center regression, as also adopted in recent VLM-based robotics work \cite{VLM_icra2024}. (ii) Alternatively, explicit coordinate-based input modulation \cite{visontransformer,dinov2} can further enhance spatial sensitivity, but these techniques typically require architectural modifications and are not generally compatible with standard few-shot VLM fine-tuning. Therefore, our baseline comparisons focus on approaches that are both practical and directly comparable within our experimental setup.

The performance of the models was evaluated by measuring the Euclidean deviation between detected and ground truth object positions, with lower deviation values indicating better accuracy. Figure \ref{fig:GridMask_results} summarizes the performance across different model groups, highlighting key findings:

\begin{figure*}[htbp]
    \centering
    \includegraphics[width=0.8\linewidth]{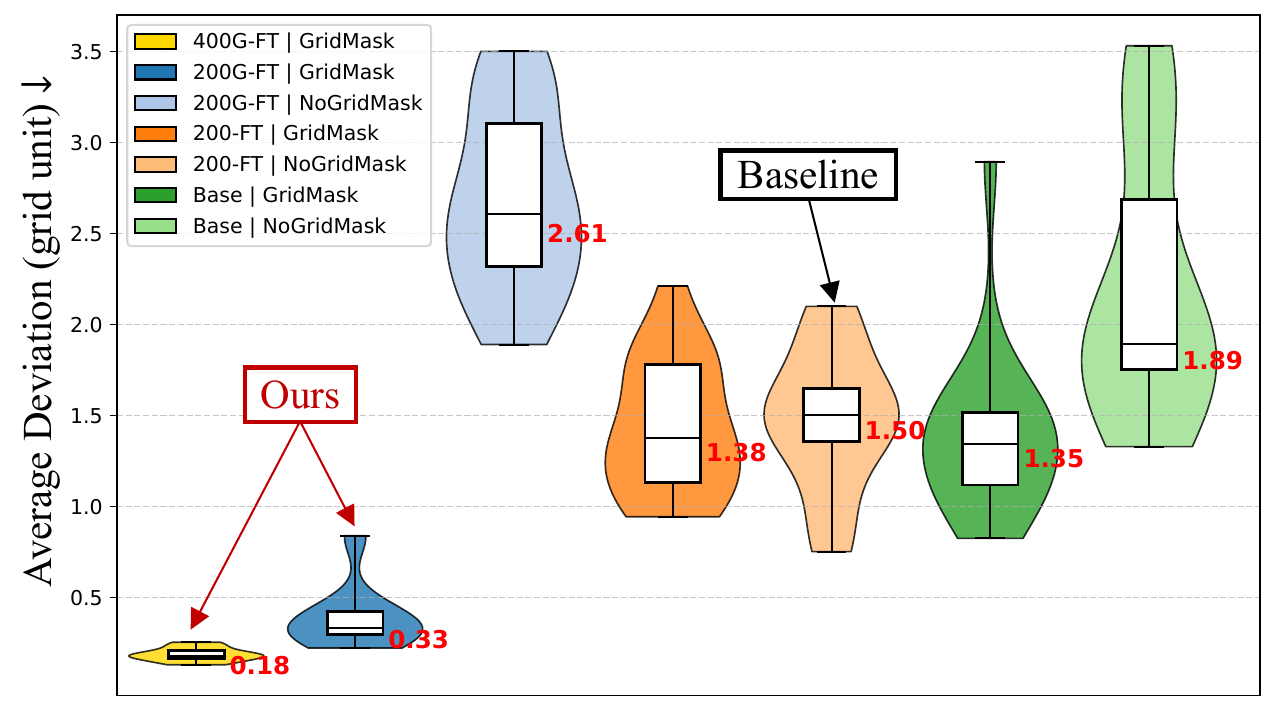}
    \caption{\textcolor{black}{Performance comparison of different models based on objects' average Euclidean deviation from ground-truth positions in the image (in grid units).}}

    \label{fig:GridMask_results}
\end{figure*}

\begin{itemize}
    \item \textcolor{black}{Models consistently perform better on images with GridMask annotations, regardless of fine-tuning. This suggests that GridMask annotations inherently improve spatial perception even in a zero-shot situation (Comparison between Base $|$ GridMask \& Base $|$ NoGridMask).}
    \item \textcolor{black}{Compared to the baseline (200-FT $|$ NoGridMask), our method (200G-FT $|$ GridMask) achieves a 78\% error reduction, substantially improving VLM’s localization ability. Additionally, the best performance, with a median deviation of around 0.15 units, is achieved by models trained and tested with GridMask annotations (400G-FT $|$ GridMask).}
    \item Models fine-tuned with GridMask data (200G-FT) exhibit weaker performance when tested on non-GridMask images compared to models fine-tuned without GridMask data (200-FT), indicating reduced generalization capability due to dependency on GridMask annotations. To mitigate this reduced generalization, several strategies could be considered, such as deploying a hybrid inference strategy (switching between GridMask-specific and standard models based on input conditions) or training with mixed datasets (combining GridMask and non-GridMask images). Nevertheless, it is important to emphasize that our GridMask-based fine-tuning approach is explicitly optimized for scenarios requiring accurate 2D spatial localization and precise semantic reasoning. For applications or tasks where high spatial precision is unnecessary, a standard model which even the base model without fine-tuning or GridMask annotations would be sufficient (which achieves an average deviation of 1.89 units).
\end{itemize}

\textcolor{black}{According to the results, fine-tuned baselines incur errors of around 1.5–2.0 units, which are sufficient for coarse navigation but prohibitive for precise manipulation.
By contrast, GridMask-enhanced models reduce errors below 0.2 units, directly enabling precise pick-and-place operations and stable aerial–ground collaboration.}

To capitalize on this improvement and further strengthen robustness, we employed a larger model (1000G-FT $|$ GridMask) in subsequent simulations and real-world experiments.

\subsection{Experiment in Simulation}
In this section, \textcolor{black}{our system follows the setup in Section~\ref{subsec:system-env}. In a simulation environment, we test whether the system generalizes in a zero-shot way to previously unseen object categories, spatial layouts, and scene complexity. We also compare low-level task completion against baseline methods.}

\subsubsection{Zero-Shot Generalization Across Novel Categories and Layouts}

\begin{figure*}[ht]
    \centering
    \subfloat[Bird-view of standard letter block scenes]{%
       \includegraphics[width=\linewidth]{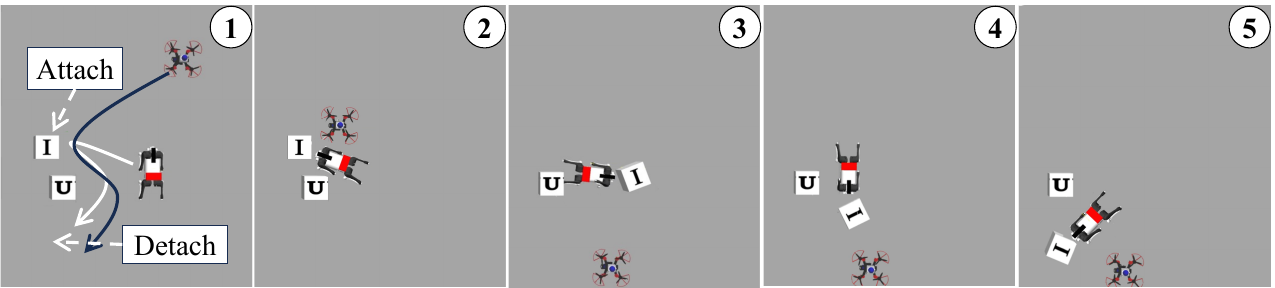}
       \label{fig:letter_sim}
    }\\
    \subfloat[Bird-view of color block scenes]{%
       \includegraphics[width=\linewidth]{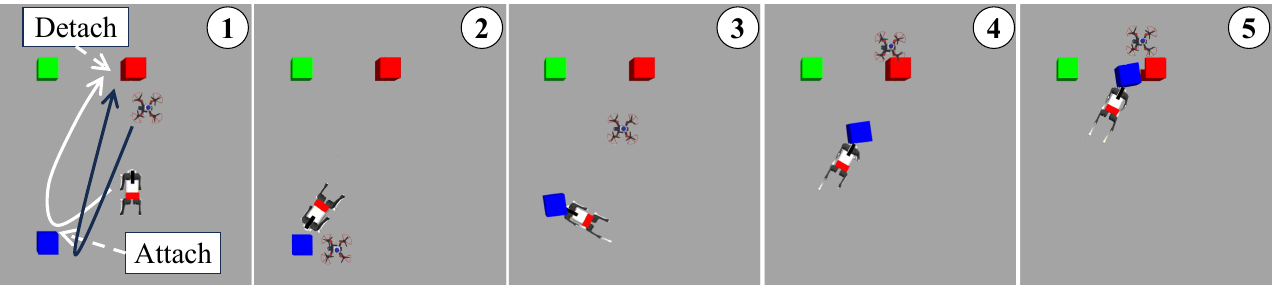}
       \label{fig:color_sim}
    }\\
    \subfloat[Bird-view of clustered mixed scene]{%
       \includegraphics[width=\linewidth]{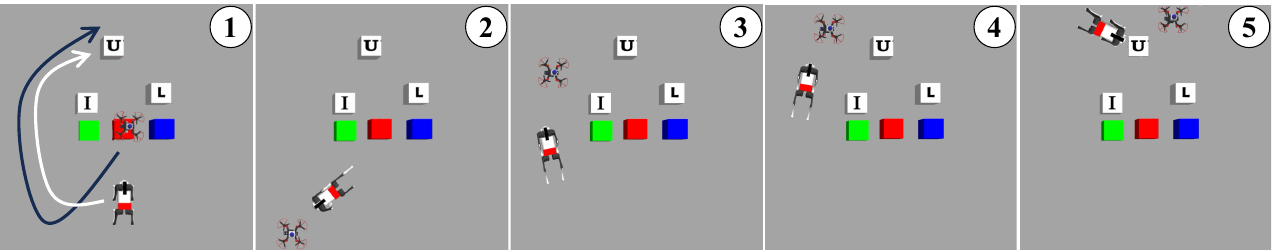}
       \label{fig:cluster_sim}
    }
    \caption{\textcolor{black}{Bird-view snapshots of simulation experiment in different scenes.} Simulation experiments demonstrating the zero-shot transferability and generalizability of the proposed system across various environments and task requirements.}
    \label{fig:sim}
\end{figure*}

To evaluate the zero-shot transferability and generalizability of our hierarchical MA-LLM framework, we first subject it to rigorous simulation-based validation across three distinct task environments of increasing semantic and perceptual diversity.

\begin{itemize}
    \item {Standard Letter Block Scene}: The workspace contains regularly arranged letter blocks, replicating the original pick-transport-place setting (Figure~\ref{fig:letter_sim}). The system was tasked to ``Move the I cube to the back side of the U cube," successfully leveraging previously learned semantic and spatial knowledge.

    \item {Color Block Scene}: The environment is populated with colored blocks, presenting novel visual categories not encountered during the fine-tuning process (Figure~\ref{fig:color_sim}). The given instruction, ``Move the blue cube to the left side of the red cube," tested the system's semantic generalization to \textcolor{black}{unseen object categories and relational instructions}. The system correctly identified and localized the colored blocks, demonstrating robust zero-shot perception and reasoning.

    \item {Clustered Mixed Scene}: This scenario combines letter blocks and colored blocks in a clustered, partially overlapping manner, significantly increasing spatial complexity and introducing semantic ambiguity (Figure~\ref{fig:cluster_sim}). The task ``Move to the U cube" required precise semantic discrimination and spatial reasoning in a challenging visual context. The system successfully navigated the ambiguity, effectively identifying and approaching the correct target object while avoiding collisions with adjacent objects.
\end{itemize}

\textcolor{black}{Across all scenes, we keep the full stack (planning, perception, execution) unchanged. The consistent success indicates the system maintains zero-shot generalization to new categories, compositions, and layouts without any simulation-specific tuning (the pipeline is only tuned for real-world trials).}

\subsubsection{\textcolor{black}{Comparative Evaluation of Planning in Low-Level Tasks}}
We further evaluate our system against representative DRL baselines on target search tasks, to evaluate low-level planning ability and highlight the trade-offs between global semantic reasoning and local reactive policies. Table~\ref{tab:conceptual_comparison} summarizes the conceptual distinctions between representative DRL approaches, recent representative VLA works and our proposed system.
While DRL methods typically focus on low-dimensional control policies trained on homogeneous agents with local egocentric observations, our framework leverages LLM–VLM modules for global, semantic-map-based planning across heterogeneous platforms. Beyond DRL approaches, the RT-series~\cite{rt2,rtx} exemplifies recent progress in end-to-end VLA learning, yet is limited to single-robot manipulation tasks.
This highlights that our contribution lies not only in performance but also in expanding the dimensionality and scope of robotic task planning.

\begin{table*}[h]
\centering
\caption{Comparison of Methodological Dimensions across DRL, RT-series, and Our Proposed System}
\label{tab:conceptual_comparison}
\begin{tabular}{lcccc}
\toprule
\textbf{Method Category} & \textbf{Task Scope} & \textbf{Input Modality} & \textbf{Planning Level} & \textbf{Adaptivity} \\
\midrule
\makecell[l]{End-to-End DRL\\\cite{xiao2024toward, chen2025target}}
& \makecell[c]{Low-level\\navigation / search}
& \makecell[c]{Local sensor\\(RGB / LiDAR)} 
& \makecell[c]{Reactive /\\short horizon} 
& \makecell[c]{Policy adaptation via\\RL training} \\
\midrule
\makecell[l]{RT-series\\\cite{rt2,rtx}} 
& \makecell[c]{Single-robot\\manipulation} 
& \makecell[c]{Vision-language\\paired embeddings} 
& \makecell[c]{Mid-level\\skill chaining} 
& \makecell[c]{Generalization via\\large-scale pretraining} \\
\midrule
\textbf{\makecell[l]{Ours\\(LLM+VLM)}} 
& \makecell[c]{Multi-agent\\heterogeneous\\task planning} 
& \makecell[c]{Natural language\\command} 
& \makecell[c]{LLM-driven\\symbolic reasoning} 
& \makecell[c]{Modular recovery +\\prompt-based adaptation} \\
\bottomrule
\end{tabular}
\end{table*}

Complementary to this methodological comparison, we also evaluate our framework against representative DRL baselines on target search tasks, in order to highlight the performance trade-offs between local reactive policies and global semantic-map reasoning.
In our experimental setting, the robot system is assigned to locate either one target (1T) or three targets (3T), which requires sequential navigation to the designated objects.
Under these conditions, our framework demonstrates a few trajectory steps and high task success rates, while maintaining robustness even as task complexity increases.

\begin{table*}[h]
\centering
\caption{Target search performance comparison between egocentric DRL baselines and our exocentric semantic-map planner. Metrics: ASR = Average Success Rate, ATS = Average Trajectory Steps, Time = average decision latency per step.}
\label{tab:comparison_egocentric_exocentric}
\begin{tabular}{
l l l
S[table-format=2.1] S[table-format=4.0]
S[table-format=2.1] S[table-format=4.0]
c
}
\toprule
\multirow{2}{*}{\textbf{Method}} & \multirow{2}{*}{\textbf{Agent Type}} & \multirow{2}{*}{\textbf{Obstacles}} &
\multicolumn{2}{c}{\textbf{1T}} & \multicolumn{2}{c}{\textbf{3T}} &
\multirow{2}{*}{\textbf{\makecell[l]{Time $\downarrow$\\(s/step)}}}\\
\cmidrule(lr){4-5}\cmidrule(lr){6-7}
& & & {\textbf{ASR $\uparrow$ (\%)}} & {\textbf{ATS $\downarrow$}} & {\textbf{ASR $\uparrow$ (\%)}} & {\textbf{ATS $\downarrow$}} & \\
\midrule
\multicolumn{8}{c}{\textbf{DRL: Egocentric sensor inputs (local perception, pre-defined target)}}\\
\midrule
POMA (1 Drone) \cite{xiao2024toward} & Homo. & 3 blocks & 43.0 & 958 & 10.0 & 1435 & \boldmath $\ll \textbf{1.0}$  \\
POMA (3 Drones) \cite{xiao2024toward} & Multi-Homo. & 3 blocks & 81.0 & 539 & 43.0 & 903  & \boldmath $\ll \textbf{1.0}$  \\
UAV\textendash AGV \cite{chen2025target} & Hetero. & \makecell[l]{Mine-tunnel\\obstacles} & \phantom{0}\num{67.6}\textsuperscript{*} & \multicolumn{1}{c}{--} & \multicolumn{1}{c}{--} & \multicolumn{1}{c}{--} & \boldmath $\ll \textbf{1.0}$  \\
\midrule
\multicolumn{8}{c}{\textbf{Ours: Exocentric semantic map (global perception, language-based target assignment)}}\\
\midrule
\textbf{\makecell[l]{Ours\\(LLM+VLM planner)}} & Hetero. & \makecell[l]{5 blocks (1T) /\\3 blocks (3T)} &
\textbf{90.0} & \textbf{9.5} &
\textbf{80.0} & \textbf{28.1} & {$\sim 3.0$} \\
\bottomrule
\end{tabular}
\vspace{3pt}\\
\raggedright\footnotesize\textsuperscript{*}\,Reported in a mine tunnel environment (length $>50$\,m); not strictly comparable in geometry to the $5{\times}5{\times}2$ workspace.
\end{table*}

Table~\ref{tab:comparison_egocentric_exocentric} contrasts DRL methods (POMA and UAV–AGV) with our hierarchical MA-LLM framework. 
Because sensing modality, agent heterogeneity, task formulation (pre-defined vs. language-assigned targets), and environment geometry differ (POMA and ours are evaluated on a $5{\times}5{\times}2$\,m workspace, whereas UAV–AGV reports results in a $>50$\,m mine tunnel), these numbers are not directly comparable. 
We therefore view the table as illustrating two distinct solution regimes rather than a head-to-head ranking: (i) low-latency, reactive policies trained end-to-end on egocentric inputs with fixed targets; and (ii) a planning-centric pipeline that reasons over a global semantic map and dynamically grounds language-specified targets across heterogeneous aerial–ground agents. 
In our setting, the system achieves fewer steps and higher task success with around 3.0 s/step decision latency, whereas DRL methods emphasize sub-second responsiveness. These complementary properties reflect distinct operating regimes: rapid local reactivity versus global semantic reasoning and coordination.

\subsection{Experiment in Real World}

\begin{figure*}[htbp]
    \centering
    \includegraphics[width=0.9\linewidth]{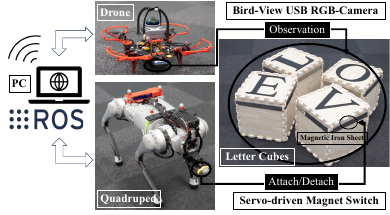}
    \caption{Hardware setup of the aerial-ground robotic system, illustrating how the drone and ground robot cooperate to pick, transport, and place target objects.}
    \label{fig:robots}
\end{figure*}

\textcolor{black}{In this section, we validate the robustness of our hierarchical MA-LLM framework under real-world conditions, using the environment setup described in Section~\ref{subsec:system-env}}. Figure~\ref{fig:robots} presents the real-world hardware setup and illustrates the interaction between the robots and the letter blocks. 

\subsubsection{\textcolor{black}{Robustness Validation}}
\begin{figure*}[h]
    \centering
    \subfloat[Dynamic disturbance during block arrangement: when a human intervenes, the system detects the suddenly appeared obstacle, enabling avoidance and successful re-attachment.]{%
       \includegraphics[width=\linewidth]{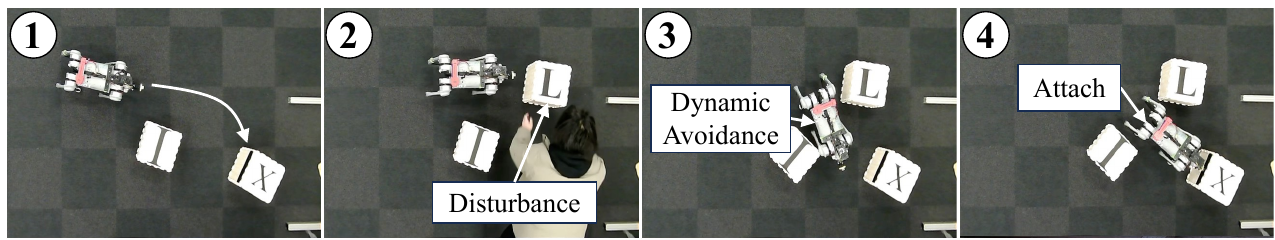}
       \label{fig:disturb1}
    }\\
    \subfloat[Manipulation disturbance: when the target block is unexpectedly removed by human, the system rolls back and retries attachment, ensuring task completion.]{%
       \includegraphics[width=\linewidth]{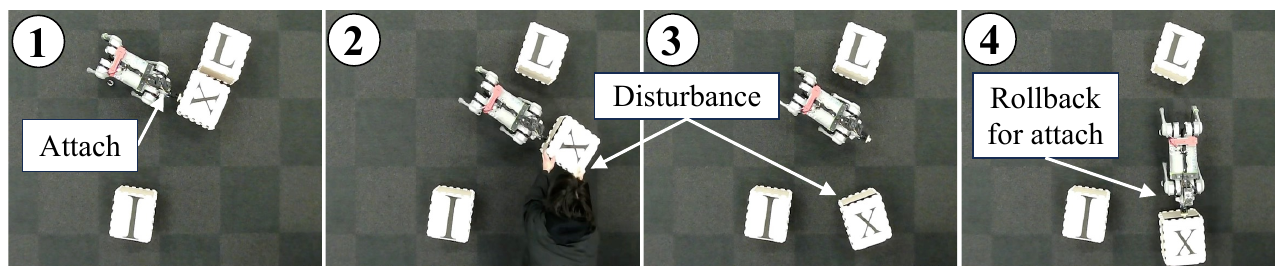}
       \label{fig:disturb2}
    }\\
    \subfloat[Perceptual variation tests: the system maintains stable reasoning and perception under different lighting (brighter/dimmer) and reduced field of view (smaller FOV).]{%
       \includegraphics[width=\linewidth]{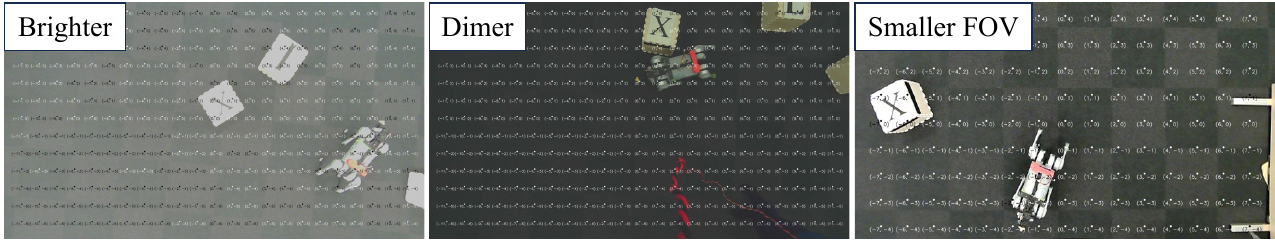}
       \label{fig:disturb3}
    }
    \caption{\textcolor{black}{Robustness of the proposed MA-LLM framework under disturbances and perceptual variations.}}
    \label{fig:disturb}
\end{figure*}

To evaluate the robustness of the proposed MA-LLM framework in challenging real-world conditions, we conducted experiments involving dynamic disturbances, manipulation interference, and perceptual variations. The scenes are summarized in Fig.~\ref{fig:disturb}.
\begin{itemize}
    \item Dynamic Disturbance: During block arrangement (Fig.~\ref{fig:disturb1}), a human deliberately introduced sudden obstacles into the robot’s trajectory. The aerial robot updated the global semantic map and the planner adjusted the ground robot’s path, enabling dynamic avoidance and subsequent attachment of the block. This demonstrates the system’s capacity to adapt to unexpected environmental changes.
    \item Manipulation Interference: In Fig.~\ref{fig:disturb2}, when the ground robot was attaching a block, the target was removed manually by a human. The system detected the failed attachment, rolled back, and retried until successful. Such recovery behavior highlights the framework’s resilience to low-level task execution errors.
    \item Perceptual Variations: As shown in Fig.~\ref{fig:disturb3}, the system was tested under brighter/dimmer lighting conditions and reduced camera field of view. Despite these variations, the VLM maintained stable recognition, leading to successful task completion. 
\end{itemize}

Overall, these results validate that the hierarchical MA-LLM framework not only supports task execution in static environment but also exhibits strong robustness against disturbances and perceptual uncertainty, which is essential for deployment in unstructured real-world environments.

\subsubsection{Real-World Application Validation}

\textcolor{black}{We further validate the hierarchical MA-LLM framework in real-world assembly tasks, focusing on its ability to decompose complex instructions and execute them through reliable perception, planning, and control.}

\begin{figure*}[htbp]
    \centering
    \subfloat[Execution of simple tasks (Type A \& B) such as ``move O cube to the right side of the K cube" or ``assembling `OK' but do not move K."]{%
       \includegraphics[width=\linewidth]{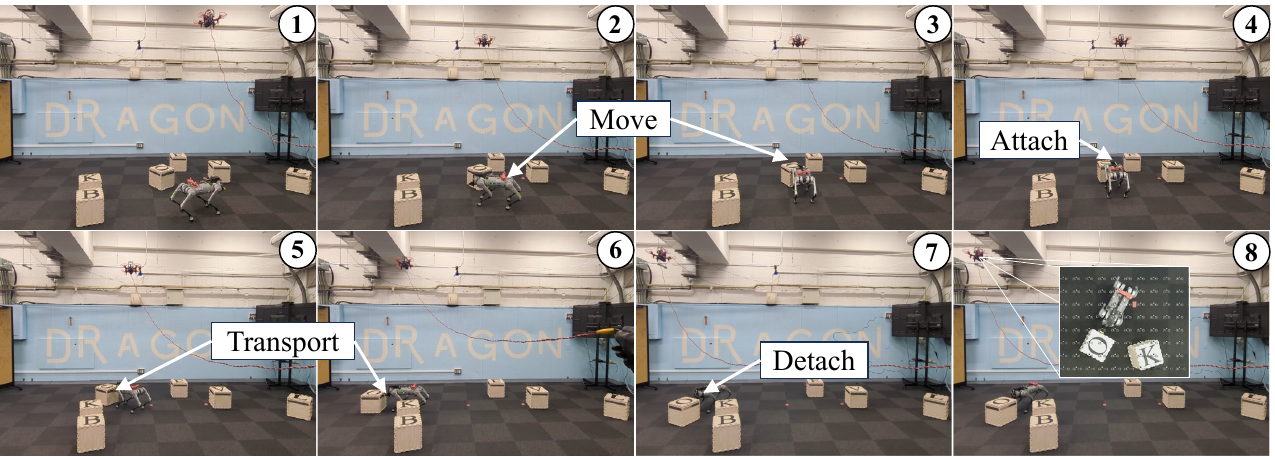}
       \label{fig:AB}
    }\\
    \subfloat[Execution of complex task (Type C) involving ``Assemble the word `LOVE'."]{%
       \includegraphics[width=\linewidth]{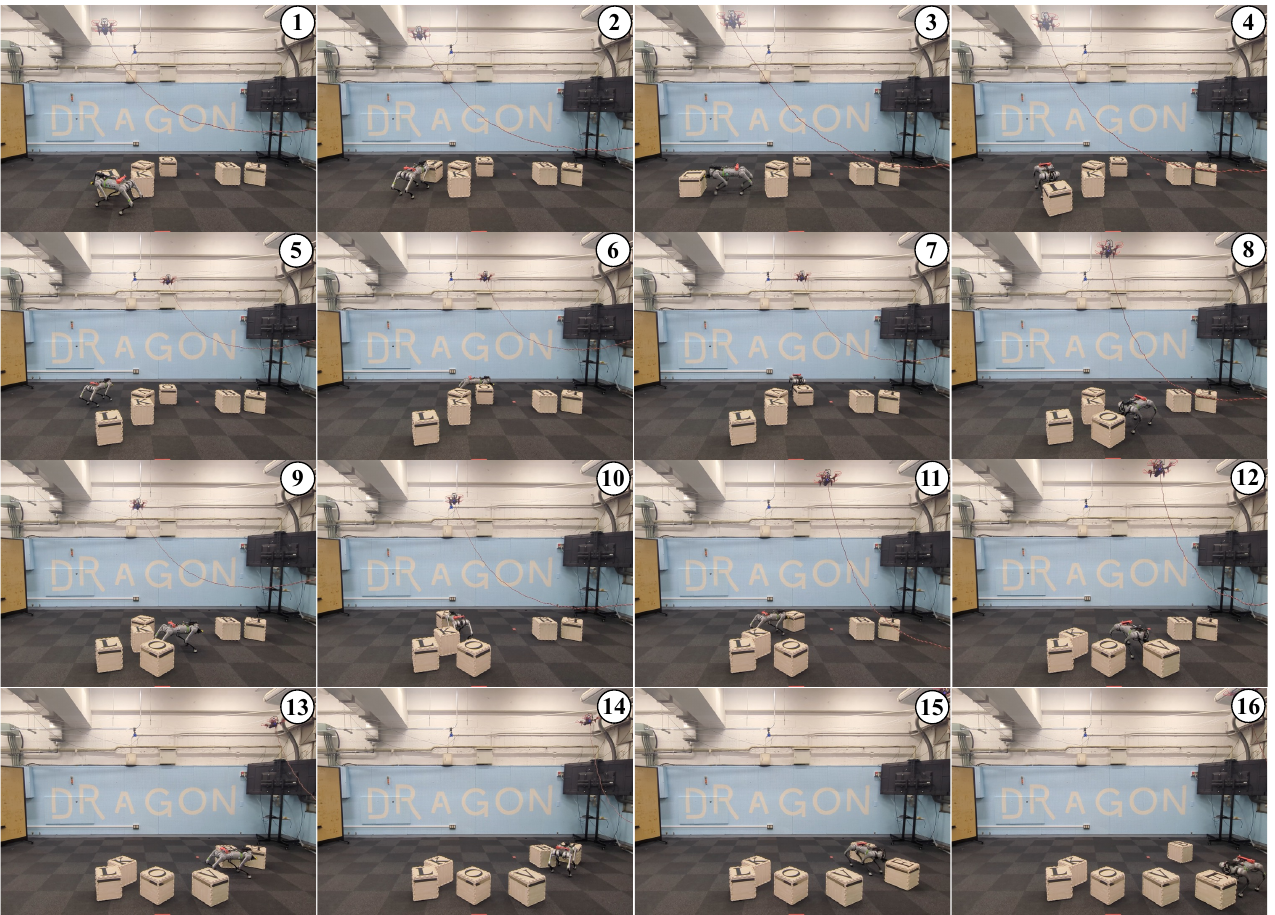}
       \label{fig:C}
    }
    \caption{Real-world experiments showing (a) Type A \& B: Simple tasks and simple tasks that require reasoning; and (b) Type C: Complex tasks that require strategic arrangement.}
    \label{fig:exp}
\end{figure*}

We categorize tasks into three types with increasing reasoning complexity: (A) direct symbolic grounding, (B) reasoning under constraints, and (C) long-horizon compositional reasoning. The details are as follows:
\begin{itemize}
    \item {Type A:} Simple tasks with direct commands, such as moving the ``L'' to the front side of the ``O'' cube.
    \item {Type B:} Simple tasks but require reasoning, such as assembling words like ``OK'' without moving one of the blocks.
    \item {Type C:} Complex, long-horizon tasks that require strategic arrangement. For example, assembling the word ``LOVE'' with flexible sequence or optional fixed cube positions.
\end{itemize}
The representative examples of these tasks are shown in Figure~\ref{fig:exp}.

In addition, we validate the VLM-based localization accuracy in real-world scenarios by comparing:  
a) the ground robot’s position error, derived from aerial robot positions obtained via SLAM;  
b) the aerial robot’s yaw is fixed to zero and aligned with the world frame, allowing the ground robot’s yaw to be directly evaluated in that frame.
\begin{figure*}[htbp]
    \centering
    \subfloat[Position error distribution of the ground robot. ]{%
       \includegraphics[width=0.4\textwidth]{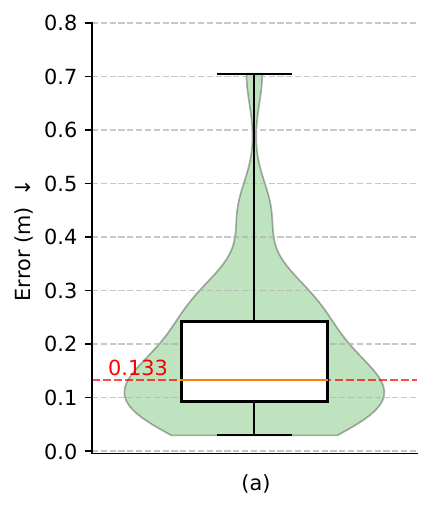}
       \label{fig:pos_error}
    }
    \hspace{1.3em}
    \subfloat[Orientation error distribution of the ground robot.]{%
       \includegraphics[width=0.4\textwidth]{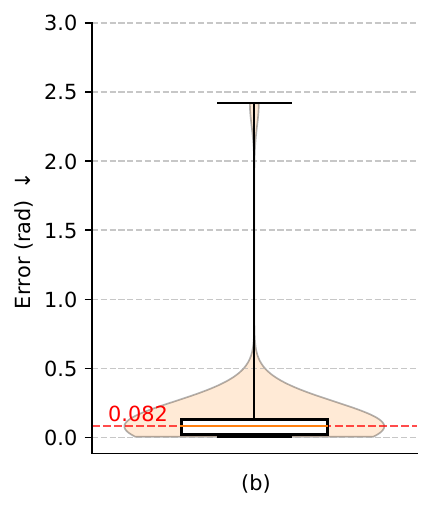}
       \label{fig:orien_error}
    }
    \caption{\textcolor{black}{Ground robot pose error distribution in the real-world experiments.}}
    \label{fig:sysvsground truth}
\end{figure*}
As shown in Figure~\ref{fig:sysvsground truth}, the median of the position error is approximately 0.13 m (Figure~\ref{fig:pos_error}), and the median of the orientation error is around 0.08 rad (Figure~\ref{fig:orien_error}). Apart from model errors, the observed deviations are primarily attributed to the aerial robot’s pose instability, which introduces transformation errors. In addition, residual lens distortion and projection inaccuracies contribute to spatial estimation error.  
These errors, while noticeable, remain sufficiently small to permit reliable pick–and–place and cooperative manipulation, as confirmed by overall task success rates. in Table~\ref{tab:eval}.

\begin{table*}[hbtp]
\centering
\caption{\textcolor{black}{Quantitative evaluation of different task types. Success rates are reported as mean ± standard deviation (Bernoulli variance). Collision rate indicates the average frequency of contact per pick–and–place trial.}}

\label{tab:eval}
\begin{tabular}{@{}l l c c c@{}}
\toprule
\hspace{2mm}\textbf{Task Type} & \textbf{Metric} & \textbf{Trials} & \multicolumn{2}{c}{\textbf{Success Rate $\uparrow$}}\\
\midrule
\hspace{2mm}\multirow{3}{*}{\makecell[l]{Type A: Simple Task\\(Direct Command)}} & Task Decomposition &  \multirow{3}{*}{5} & \multicolumn{2}{c}{1.00}\\
& Semantic Map Construction & & \multicolumn{2}{c}{0.80 $\pm$ 0.18}\\
& Task Completion & & \multicolumn{2}{c}{0.80 $\pm$ 0.18}\\
\midrule

\hspace{2mm}\multirow{3}{*}{\makecell[l]{Type B: Simple Task\\(Reasoning Required)}} & Task Decomposition & \multirow{3}{*}{5} & \multicolumn{2}{c}{0.80 $\pm$ 0.18}\\
& Semantic Map Construction & & \multicolumn{2}{c}{1.00}\\
& Task Completion &  & \multicolumn{2}{c}{0.80 $\pm$ 0.18}\\
\midrule
\hspace{2mm}\multirow{3}{*}{\makecell[l]{Type C: Complex Task\\(Long-Horizon Reasoning)}} & Task Decomposition & \multirow{3}{*}{5} & \multicolumn{2}{c}{1.00}\\
& Semantic Map Construction & & \multicolumn{2}{c}{0.80 $\pm$ 0.18}\\
& Task Completion &  & \multicolumn{2}{c}{0.80 $\pm$ 0.18}\\
\midrule
\multicolumn{2}{c}{\textbf{Overall Task Completion (Type A+B+C)}} & 15 &  \multicolumn{2}{c}{\textbf{0.80 $\pm$ 0.10}}\\
\toprule
\multicolumn{2}{c}{\textbf{Failure Source}} 
  & \textbf{Total Trials} & \multicolumn{2}{c}{\textbf{Failure Rate $\downarrow$}} \\
\midrule
\multicolumn{2}{c}{Task Planning}            & \multirow{3}{*}{15} & \multicolumn{2}{c}{0.07 $\pm$ 0.06} \\
\multicolumn{2}{c}{Semantic Map} &                      & \multicolumn{2}{c}{0.13 $\pm$ 0.09} \\
\multicolumn{2}{c}{Behavior in Execution}                 &                      & \multicolumn{2}{c}{0.00} \\
\toprule
\multicolumn{2}{c}{\textbf{Common Metrics}} & \textbf{Total Trials} & \textbf{\makecell[c]{Accuracy $\uparrow$}} & \textbf{\makecell[c]{Collision Rate $\downarrow$}}\\
\midrule
\multicolumn{2}{c}{Collisions} & 25 & \textemdash & 0.68 $\pm$ 0.09 \\
\multicolumn{2}{c}{Recognition Accuracy} & \multirow{2}{*}{378} & 0.74 $\pm$ 0.02 & \textemdash\\
\multicolumn{2}{c}{Semantic Labeling Accuracy} &  & 1.00 $\pm$ 0.00 & \textemdash\\
\bottomrule
\end{tabular}
\end{table*}

Furthermore, the real-world applicability of the system was evaluated using six key metrics, as summarized in Table~\ref{tab:eval}:

\begin{itemize}

    \item Task Decomposition Accuracy: Evaluates the correctness of translating user instructions into robot-specific sub-tasks. Generally, the system demonstrated strong decomposition capabilities. However, failures occurred once with commands involving negation of spatial constraints, such as misinterpreting the spatial relation in the instruction \textit{“assemble the word BE but do not move B”}, incorrectly placing the E cube on the left instead of the right.

    \textcolor{black}{\item Semantic Map Construction: Evaluates the accuracy (87\%) of merging local semantic maps into a consistent global map. The system effectively corrected overlapping labeling errors; for instance, when one local map misclassified an object but others provided the correct label, the LLM reconciled the discrepancy. However, it struggled with errors in isolated local maps. For example, when a letter “L” was misrecognized as “I” and no other local map observed the same region, leaving the system without cross-validation to correct the mistake. This illustrates its reliance on redundancy across local observations to maintain robustness.}
    
    \item \textcolor{black}{Task Completion: Across all task types (A–C), the system achieved an overall success rate of 80\% (15 trials), demonstrating reliable robustness in executing complex multi-step tasks.}

    \item \textcolor{black}{Failure Source: Among the three observed failures, two originated from inaccuracies in the global semantic map construction that impaired aerial path planning, and one stemmed from initial decomposition errors. No failures occurred at the execution layer, confirming that low-level controllers remained stable once valid plans were provided. These results indicate that the primary but rare sources of error lie in high-level reasoning and semantic perception rather than motion control.}
    
    \item Collisions: Records the frequency of collisions, averaging 0.68 per pick–transport–place operation. This number also accounts for minor contacts that did not affect task success. Quantitative error analysis (Figure~\ref{fig:sysvsground truth}) shows that most collisions stem from small deviations in estimated object poses and inherent limits of motion control precision. In cluttered scenarios, such minor contacts are occasionally unavoidable, as the planner prioritizes an optimized global path that balances collision avoidance with efficiency. Importantly, these collisions never prevented task completion, confirming the robustness of the system in semantic navigation and manipulation tasks.
    
    \textcolor{black}{\item Recognition Accuracy and Semantic Labeling Accuracy: Visual recognition accuracy reached 74\%, with most errors due to rotational ambiguity between visually similar cubes. Crucially, once recognition was correct, semantic labeling was perfectly accurate across all 378 trials, ensuring that roles such as \texttt{main}, \texttt{target}, \texttt{landmark}, and \texttt{obstacle} were consistently and reliably assigned.}
    
\end{itemize}

Overall, these real-world experiments demonstrate that the hierarchical MA-LLM framework robustly integrates perception, reasoning, and execution. \textcolor{black}{Despite occasional perception errors and semantic ambiguities, the system consistently decomposes complex instructions and executes multi-step assembly tasks, highlighting its potential for deployment in real-world collaborative robotics.}

\section{Conclusion}
This paper presents a hierarchical MA-LLM framework that integrates LLM-based task reasoning, VLM-based semantic perception, and motion-level execution for an aerial–ground heterogeneous robotic system. Through a structured three-layer design, the system interprets natural language commands, constructs semantic maps, and executes complex manipulation tasks via coordinated aerial–ground behaviors. \textcolor{black}{Unlike prior frameworks (e.g., COHERENT \cite{coherent}, RT-series \cite{rt2,rtx}), which either lack integrated perception or struggle with multi-stage reasoning, our approach bridges this gap by combining fine-tuned VLM perception with LLM-based hierarchical reasoning, enabling both closed-loop collaboration and advanced task decomposition.}

\textcolor{black}{Experimental results validate the generalizability and robustness of our framework across simulation and real-world experiments in diverse tasks. Moreover, the ablation studies further confirm the effectiveness of GridMask-based fine-tuning in enhancing spatial precision, which is critical for reliable manipulation.}

A current limitation is that in extremely cluttered environments, perception inaccuracies and optimization-based planning may jointly result in suboptimal trajectories or increased collisions. \textcolor{black}{Another practical limitation is the inference latency of large models, which is approximately 3s per decision step in our current API-based implementation. However, since this delay only affects top-level decision-making while low-level control and stabilization are executed onboard, the robot safely maintains its state during waiting periods without disruptive effects on task execution. }

\textcolor{black}{Future work could address these challenges through (i) refining perception, for instance by incorporating segmentation-enhanced recognition; (ii) extending to 3D perception for aerial robots and 3D motion planning for ground or hybrid robots; (iii) reducing inference latency through local deployment and lighter-weight variants;
and (iv) complementing the hierarchical pipeline with end-to-end VLA models, such as diffusion transformer–based planners, to combine high-level reasoning with adaptive low-level execution, thereby reducing reliance on pre-scripted motion functions.}

\textcolor{black}{In summary, this work demonstrates that combining LLM-driven reasoning with GridMask-enhanced VLM perception provides a viable pathway toward heterogeneous multi-robot collaboration. By bridging high-level semantic reasoning with precise low-level execution, our framework takes a step toward more generalizable, robust, and adaptable robotic intelligence for real-world deployment.}

%
\appendix
\subsection{\textcolor{black}{Appendix - Mathematical Details of Global Path Optimization}}
\label{app:globalpath}

\subsubsection{\textcolor{black}{Details of Computation}}
As introduced in Section \ref{subsec:globalplan}, the global path is represented by a B-spline 
$\mathbf{S}(u;\mathbf{C})$ parameterized by the control points 
$\mathbf{C} = \{\mathbf{c}_0, \ldots, \mathbf{c}_n\}$.
The global path cost function is defined as
\begin{equation}
J_\text{global}(\mathbf{S}(u;\mathbf{C}), \mathbf{O}) 
= Q_L L(\mathbf{S}) + Q_K K(\mathbf{S}) + Q_{H}H(\mathbf{S}, \mathbf{O}),
\end{equation}
where the three components are:
\begin{itemize}
    \item \( L(\mathbf{S}) \): Path length cost, penalizing excessively long paths.
    \begin{equation}
    L(\mathbf{S}) = \sum_{i=1}^{N-1} \| \mathbf{S}(u_i;\mathbf{C}) - \mathbf{S}(u_{i-1};\mathbf{C}) \|.
    \end{equation}
    \item \( K(\mathbf{S}) \): Path curvature cost, promoting smoothness.
    \begin{equation}
    K(\mathbf{S}) = \sum_{i=2}^{N-2} \| \mathbf{S}(u_{i+1};\mathbf{C}) - 2\mathbf{S}(u_i;\mathbf{C}) + \mathbf{S}(u_{i-1};\mathbf{C}) \|.
    \end{equation}
    \item \( H(\mathbf{S}, \mathbf{O}) \): Obstacle avoidance cost. For each obstacle 
    \( \mathbf{o}_j \in \mathbf{O} \) ($j=0,\dots,m-1$) 
    and each sampled point \( \mathbf{S}(u_i;\mathbf{C}) \), a penalty is applied if the distance is less than \( d_{\text{safe}} \):
    \begin{equation}
    H(\mathbf{S}, \mathbf{O}) =  \sum_{i=0}^{N-1}\sum_{j=0}^{m-1} 
    \max\!\left(0, d_{\text{safe}} - \| \mathbf{S}(u_i;\mathbf{C}) - \mathbf{o}_j \| \right)^2.
    \end{equation}
\end{itemize}
Here, \( Q_L, Q_K, Q_{H} \) are weighting coefficients balancing the respective terms. \( N \) is the number of samples, m is the number of obstacles.

\subsubsection{\textcolor{black}{Computational Complexity}}
The evaluation of the global path cost involves three components.
The length and curvature terms each require $O(N)$ operations over $N$ sampled path points,
while the obstacle avoidance term requires $O(mN)$ operations, where $m$ is the number of obstacles.
Hence, the overall complexity of a single cost evaluation is $O(mN)$.

In practice, both $N$ and $m$ are small ($N < 50$, $m \leq 6$ in our experiments),
making the computation lightweight. Even with $k$ optimization iterations,
the resulting complexity $O(kmN)$ remains tractable.
On a standard desktop CPU (3–5 GHz, single thread), the computation typically completes within 0.5–2 ms per iteration with a vectorized implementation,
which is sufficient to support real-time optimization at hundreds of hertz.

\subsection{\textcolor{black}{Appendix - Mathematical Details of Local Velocity Direction Optimization}}
\label{app:localdirection}
\subsubsection{\textcolor{black}{Details of Computation}}
As introduced in Section \ref{subsec:localplan}, the local planner selects the optimal velocity direction 
\(\theta^*\) from candidate velocity directions $\theta \in \{\theta_0, \theta_1, \ldots, \theta_{n-1}\}$ by minimizing a weighted cost function. The local velocity direction cost function is given by
\begin{align}
    J_{\text{local}}(\theta,\mathbf{M},\mathbf{T},\mathbf{O}) =\ & Q_{A} A(\theta,\mathbf{M},\mathbf{T}) + Q_{Z} Z(\theta,\mathbf{M})\nonumber \\& + Q_{H} H(\theta,\mathbf{O}) + Q_{W} W(\theta),
\end{align}
where each term is defined as follows:
\begin{itemize}
    \item \(A(\theta,\mathbf{M},\mathbf{T})\): This term measures the deviation between the candidate direction \( \theta \) and the target direction. 
    \begin{equation}
        A(\theta,\mathbf{M},\mathbf{T}) = \frac{\beta}{\|\mathbf{T} - \mathbf{M}\|}  \Delta\theta,\quad \Delta\theta = \arccos\big({d}_{\theta} \cdot {d}_{\text{goal}}),
    \end{equation}
  where \( {d}_{\theta} = [\cos(\theta), \sin(\theta)]^T \) is the unit vector in direction \( \theta \), and \({d}_{\text{goal}} = \frac{\mathbf{T-M}}{\|\mathbf{T} - \mathbf{M}\|}\) is the normalized vector from the \texttt{main} coordinate \( \mathbf{M} \) to the \texttt{target} coordinate \( \mathbf{T} \). The bias factor \( \frac{\beta}{\|\mathbf{T} - \mathbf{M}\|} \) adjusts the sensitivity to angular deviation based on the distance to the target.
  \item \(Z(\theta,\mathbf{M})\): This term encourages the result to tend to the zero point in the image.
        \begin{equation}
            Z(\theta,\mathbf{M}) = \Delta\theta_\text{zero},\quad \Delta\theta_\text{zero} = \arccos\big({d}_{\theta} \cdot {d}_{\text{zero}}\big),
        \end{equation}
    where \({d}_{\text{zero}} = \frac{\mathbf{Z} - \mathbf{M}}{\|\mathbf{Z} - \mathbf{M}\|}\) is the normalized vector from the \texttt{main} \( \mathbf{M} \) to the zero point \( \mathbf{Z} \).
  \item \(H(\theta,\mathbf{O})\): For a given candidate direction, this term evaluates the proximity to each \texttt{obstacle} 
    \( \mathbf{o}_j \in \mathbf{O} \) ($j=0,\dots,m-1$) along the direction.
  \begin{equation}
      H(\theta,\mathbf{O}) = \sum_{j=0}^{m-1} 
    \begin{cases}
    \displaystyle \frac{1}{d_\perp(\mathbf{o}_j, \theta) + \epsilon},  
    &\text{if } d_\perp(\mathbf{o}_j, \theta) < d_{\text{safe}}, \\[1mm]
    0, & \text{otherwise},
    \end{cases}
  \end{equation}
  where $d_\perp(\mathbf{o}_j, \theta)$ denotes the perpendicular distance from obstacle $\mathbf{o}_j$ to the candidate ray along direction $\theta$. \( d_{\text{safe}} \) is a safety distance threshold, \( \epsilon \) is a small constant to prevent division by zero, and $n$ is candidate directions.
  \item \(W(\theta)\): This term enforces the boundary conditions to ensure that the ground robot's next advance along \( \theta_i \) remains within the predefined workspace window of the semantic map.
 \begin{equation}
      W(\theta) =
      \begin{cases}
      \infty, & \text{if exceeding boundaries}, \\
      0, & \text{otherwise}.
      \end{cases}
 \end{equation}   
\item  \( Q_A \), \(Q_Z\), \( Q_H \), and \( Q_W \) are weighting coefficients balancing the respective components. m is the number of obstacles.
\end{itemize}

\subsubsection{\textcolor{black}{Computational Complexity}}
The evaluation of the local velocity cost requires $O(1)$ operations per candidate for target deviation, zero-point alignment, and boundary constraints, resulting in $O(n)$ operations across $n$ candidate directions.
The obstacle avoidance term requires $O(m)$ operations, where $m$ is the number of obstacles. 
Thus, with $n$ candidate directions, the total complexity is $O(nm)$. 
In our implementation, $n=360$ (1° resolution over $360^\circ$) and $m \leq 6$, 
so the runtime per optimization cycle remains in the millisecond range. On a standard desktop CPU (3–5 GHz, single thread), this corresponds to approximately 0.2–0.8 ms per cycle with a vectorized implementation, enabling planning rates well above 1 kHz.

\bibliographystyle{IEEEtran}
\bibliography{citation}



\end{document}